%% file: main.tex
\pdfoutput=1

\documentclass[11pt]{article}

\usepackage[]{EMNLP2023}
\usepackage{graphicx}
\usepackage[export]{adjustbox}
\usepackage{caption}
\usepackage{subcaption}
\usepackage{amsmath}
\usepackage{times}
\usepackage{latexsym}
\usepackage{hyperref}       
\usepackage{url}
\usepackage{xcolor, soul}
\usepackage{graphicx}
\usepackage{booktabs}
\usepackage{multirow}
\usepackage{arabtex}
\usepackage{utf8}
\usepackage{pdflscape}
\setcode{utf8}
\usepackage{array}
\newcolumntype{H}{>{\setbox0=\hbox\bgroup}c<{\egroup}@{}}
\hypersetup{
colorlinks = true,
linkcolor = blue,
}

\usepackage[T1]{fontenc}

\usepackage[utf8]{inputenc}

\usepackage{microtype}

\setcode{utf8}

\title{Beyond English: \\Evaluating LLMs for Arabic Grammatical Error Correction}

\author{\normalsize Sang Yun Kwon$^{\xi}$ ~ Gagan Bhatia $^{\xi}$ ~ El Moatez Billah Nagoudi$^{\xi}$ 
~\\ {\bf Muhammad Abdul-Mageed}$^{\xi,\omega}$\\
\normalsize $^{\xi}$Deep Learning \& Natural Language Processing Group,
  The University of British Columbia\\\normalsize  $^{\omega}$Department of Natural Language Processing \& Department of Machine Learning, MBZUAI\\ %
  \texttt{\normalsize \{skwon01@student.,gagan30@student.,muhammad.mageed@\}ubc.ca}}

\begin{document}
\maketitle
\section*{~~~~~~~~~~~~~~~~~~~~~~~~~~~Abstract}
Large language models (LLMs) finetuned to follow human instruction have recently exhibited significant capabilities in various English NLP tasks. However, their performance in grammatical error correction (GEC), especially on languages other than English, remains significantly unexplored. In this work, we evaluate the abilities of instruction finetuned LLMs in Arabic GEC, a complex task due to Arabic's rich morphology. Our findings suggest that various prompting methods, coupled with (in-context) few-shot learning, demonstrate considerable effectiveness, with GPT-4 achieving up to $65.49$ F\textsubscript{1} score under expert prompting (approximately $5$ points higher than our established baseline). Despite these positive results, we find that instruction finetuned models, regardless of their size, are still outperformed by fully finetuned ones, even if they are significantly smaller in size. This disparity highlights substantial room for improvements for LLMs. Inspired by methods used in low-resource machine translation, we also develop a method exploiting synthetic data that significantly outperforms previous models on two standard Arabic benchmarks. Our best model achieves a new SOTA on Arabic GEC, with $73.29$ and $73.26$ F$\textsubscript{1}$ on the 2014 and 2015 QALB datasets, respectively, compared to peer-reviewed published baselines. 

\input{Introduction}

\input{Related_Works}

\input{Datasets_Evaluation_Metric}


\input{Setups}

\input{Results_and_Discussion}

\input{Conclusion}

\input{ack}


\bibliography{anthology,custom,citations}
\bibliographystyle{acl_natbib}
\newpage
\appendix
\input{Appendix}

\end{document}

%% file: Introduction.tex
\section{Introduction}\label{intro}
\begin{figure}[t]
  \centering
  \includegraphics[width=\linewidth]{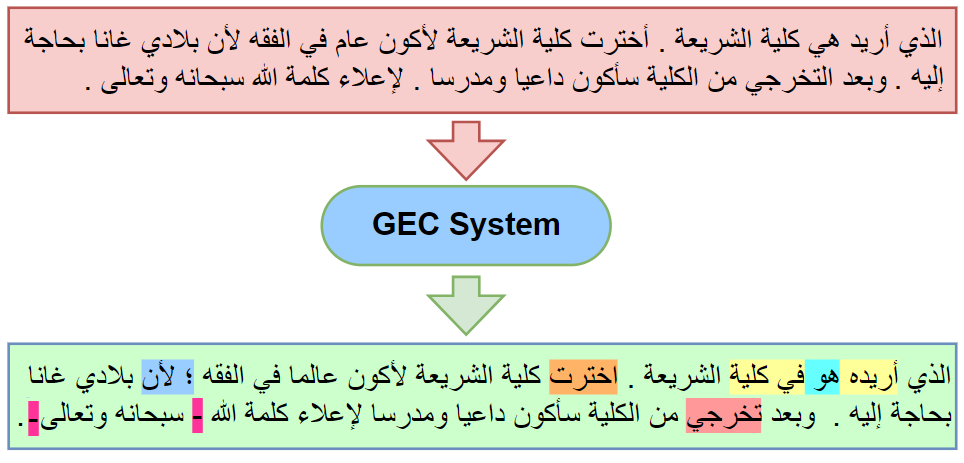}
\caption{An example of an Arabic GEC system showcasing six types of errors: \colorbox{cyan!30}{\textit{character replacement}}, \colorbox{yellow!50}{\textit{missing word}}, \colorbox{orange!35}{\textit{hamza error}}, \colorbox{blue!25}{\textit{missing punctuation}}, \colorbox{pink!55}{\textit{additional character}}, and \colorbox{violet!40}{\textit{punctuation confusion}}.}
\label{fig:trj_overview} 
\end{figure}

As interest in second language learning continues to grow, ensuring the accuracy and effectiveness of written language becomes increasingly significant for pedagogical tools and language evaluation~\cite{rothe-etal-2021-simple, tarnavskyi-etal-2022-ensembling}. A key component in this respect is grammatical error correction (GEC), a sub-area of natural language generation (NLG), which analyzes written text to automatically detect and correct diverse grammatical errors. Figure \ref{fig:trj_overview} shows an instance of GEC from~\citet{mohit2014first}. Despite the growing attention to GEC, it is predominantly studied within the English language. Extending GEC systems to other languages presents significant challenge, due to lack of high-quality parallel data and/or inherent challenges in these languages. Recognizing this, our work focuses on Arabic. In addition to being less-explored for GEC~\cite{mohit2014first, rozovskaya2015second,mohit2014first, rozovskaya2015second,
solyman2022automatic,alhafni2023advancements}, Arabic has complex grammar and rich morphology that present significant challenges and further motivate our work.

Focusing primarily on English, the field of GEC has witnessed significant advancements, specifically with the emergence of sequence-to-sequence (seq2seq)~\cite{Chollampatt_Ng_2018,gong-etal-2022-revisiting} and sequence-to-edit approaches (seq2edit)~\cite{awasthi-etal-2019-parallel,omelianchuk2020gector} achieving SoTA results in the CONLL-2014~\cite{ng-etal-2014-conll} and the BEA-2019 shared task~\cite{bryant-etal-2019-bea}, respectively. In spite of the efficacy of these approaches, they rely heavily on large amounts of labeled data. This poses issues in low-resource scenarios~\cite{feng-etal-2021-survey}. Yet, scaled up language models, \textit{aka} large language models (LLMs) have recently demonstrated remarkable potential in various NLP tasks. The core strength of LLMs lies in their capacity to generalize across a wide range of languages and tasks, and in-context learning (ICL), enabling them to handle various NLP tasks with just a few examples (i.e., few-shot learning). A key strategy for LLMs is \textit{instruction fine-tuning}, where they are refined on a collection of tasks formulated as instructions~\cite{wei2022finetuned}. This process amplifies the models' ability to respond accurately to directives, reducing the need for few-shot examples~\cite{ouyang2022training, wei2022chain, sanh2021multitask}. 

Given the ability of LLMs to adeptly address the low-resource challenge, we investigate them in the context of GEC. Focusing primarily on ChatGPT, we examine the effectiveness of various prompting strategies such as few-shot chain of thought (CoT) prompting~\cite{kojima2022large} and expert prompting~\cite{xu2023expertprompting}. Our research extends the realm of GEC research by concentrating on the unique challenges posed by Arabic. 
Drawing upon the work of \citet{junczys-dowmunt-etal-2018-approaching}, we frame these challenges within the context of a low-resource MT task. We then carefully conduct a thorough comparison of the different methodologies employed in addressing GEC in Arabic. Our key contributions in this paper are as follows:

\begin{enumerate}

\item[\textbf{1.}] We conduct a comprehensive investigation of the potential of LLMs for tasks involving GEC in Arabic.
\item[\textbf{2.}] We methodically investigate the utility of different prompting methods for generating synthetic data with ChatGPT for GEC.
\item[\textbf{3.}] We further carry out in-depth comparisons between several approaches (seq2seq, seq2edit, and instruction fine-tuning) for Arabic GEC (AGEC), allowing us to offer novel insights as to the utility of these  approaches.


\end{enumerate}







The rest of this paper is organized as follows:
In Section~\ref{RW}, we review related work with a particular emphasis on Arabic. In Section~\ref{DE}, we outline our experimental setups. We present our experiments on LLMs and prompting strategies in Section~\ref{LLM}. In Section~\ref{DA}, we introduce our seq2seq approach along with data augmentation techniques; Section~\ref{ST} discusses our seq2edit approach. In Section~\ref{AET}, we conduct a comprehensive analysis of our best model. We discuss our results in Section \ref{D}, and conclude in Section~\ref{C}.

%% file: Related_Works.tex
\section{Related Work}\label{RW}
\noindent{\bf Progress in GEC.}
Pretrained Transformer models have reframed GEC as an MT task, achieving SoTA results~\cite{ng-etal-2014-conll, felice-etal-2014-grammatical, junczys2018approaching,grundkiewicz-etal-2019-neural}. In contrast, sequence2edit approaches view the task as text-to-edit, converting input sentences into edit operations to produce corrected sentences~\cite{malmi2019encode,awasthi-etal-2019-parallel, omelianchuk2020gector}. These approaches both streamline the training process and enhance model accuracy. Further progress has also been made through methods such as instruction fine-tuning~\cite{chung2022scaling} and innovative prompting techniques, such as CoT~\cite{kojima2022large} and Expert~\cite{xu2023expertprompting} prompting. Recent applications of LLMs, like ChatGPT in GEC, highlight their potential. We provide further details on each of these methods  in Appendix~\ref{app:RW_approaches}.

\noindent{\bf Arabic GEC.}
Challenges in AGEC stem from the complexity and morphological richness of Arabic. Arabic, being a collection of a diverse array of languages and dialectal varieties with Modern Standard Arabic (MSA) as a contemporary variety, is further complicated by the optional use of diacritics. This introduces orthographic ambiguity, further complicating GEC in Arabic~\cite{abdul-mageed-etal-2020-nadi, belkebir2021automatic}. Despite these challenges, progress in AGEC has been made. This includes development of benchmark datasets through the QALB-2014 and 2015 shared tasks~\cite{mohit2014first, rozovskaya-etal-2015-second, habash-palfreyman-2022-zaebuc}, and introduction of synthetic datasets~\cite{SOLYMAN2021303, SOLYMAN2023101572}. As for model development, character-level seq2seq models~\cite{watson2018utilizing} and other novel approaches are shown to be effective on AGEC L1 data. Further details about  progress in AGEC are provided in Appendix~\ref{app:RW_approaches}. Despite this progress, no exploration has been undertaken into the utility of using ChatGPT (or other LLMs) for AGEC. Moreover, substantial work remains in exploring synthetic data generation, including the use of LLMs and the adoption of diverse machine learning approaches. Our research aims to address these gap.

%% file: Datasets_Evaluation_Metric.tex
\section{Experimental Setup}\label{DE}

\subsection{Datasets}
In this study, we make use of the QALB-2014~\cite{mohit2014first} and 2015~\cite{rozovskaya-etal-2015-second} datasets to evaluate the performance of our models. Both datasets make use of the QALB corpus~\cite{zaghouani-etal-2014-large}, a manually corrected collection of Arabic texts. These texts include online commentaries from Aljazeera articles in MSA by L1 native speakers, as well as texts produced by L2 learners of Arabic. Both the QALB 2014 and 2015 datasets are split into training (Train), development (Dev), and test (Test) sets based on their annotated dates. 
QALB 2015 includes L1 commentaries and L2 texts that cover different genres and error types. For the purposes of our study, we exclusively use the L1 test set (2015), as we focus on sentence-level AGEC, where L2 test sets are document-level. 
We used Train, Dev, and Test splits described in Table~\ref{tab:DatasetDescriptions}.

\input{Tables/Qalb_overall}




\subsection{Evaluation}
\noindent{\textbf{Metrics.}} For evaluation, we utilize the overlap-based metric MaxMatch (M$^2$)~\cite{dahlmeier-ng-2012-better}, which aligns source and hypothesis sentences based on Levenshtein distance , selecting maximal matching edits, scoring the precision (P), recall (R), and F\textsubscript{1} measure. Moreover, we report the F\textsubscript{0.5} score , a variation of the F\textsubscript{1} score that places twice as much weight on precision than on recall. This reflects a consensus, in alignment with recent works on GEC, that precision holds greater importance than error correction in GEC systems. Importantly, we use the exact scripts provided from the shared task for evaluation, ensuring consistency with other studies.





\subsection{Models \& Fine-tuning}
\noindent{\textbf{LLMs.}}
To evaluate the capabilities of LLMs for AGEC, we prompt and fine-tune LLMs of varying sizes, including LLaMA-7B \cite{touvron2023llama}, Vicuna-13B \cite{vicuna2023}, Bactrian-X$_{\textit{bloom}}$-7B~\cite{bactrian}, and Bactrian-X$_{\textit{llama}}$-7B~\cite{bactrian}. For experiments with ChatGPT, we use the official API to prompt ChatGPT-3.5 Turbo and GPT-4. We instruction fine-tune each smaller model for $4$ epochs using a learning rate of 2e-5 and a batch size of $4$. We then pick the best-performing model on our Dev, then report on our blind Test.

\noindent{\textbf{Seq2seq models.}}
Our baseline settings for seq2seq models include AraBart~\cite{eddine2022arabart} and AraT5\textsubscript{v2}~\cite{nagoudi-etal-2022-arat5}, both of which are text-to-text transformers specifically tailored for Arabic. We also evaluate the performance of the mT0~\cite{muennighoff2022crosslingual} and mT5~\cite{xue2020mt5} variants of the T5 model~\cite{raffel2020exploring}, both configured for multilingual tasks. Each model is fine-tuned for $50$ epochs, with an early stopping patience of $5$ using a learning rate of $5$e-$5$ and a batch size of $32$. These models serve as the baseline for comparison throughout our experiments.

\noindent{\textbf{Seq2edit models.}} 
ARBERT\textsubscript{v2} and MARBERT\textsubscript{v2}~\cite{abdul-mageed-etal-2021-arbert} serve as the baselines for our seq2edit experiments. We fine-tune each model for $100$ epochs for each training stage, employing a learning rate of $1$e-$5$ and a batch size of $4$, with an early stopping patience of $5$.

All models are trained for three runs, with seeds of $22$, $32$, and $42$. We then select the best-performing model based on our Dev data for blind-testing on the Test sets. \textit{We report the mean score of the three runs, along with its standard deviation.} Results on the Dev set, and more details regarding hyperparameters are provided in Appendix~\ref{tab: dev}, and Appendix~\ref{tab:hyperparameters_all_models}.

%% file: Tables/Qalb_overall.tex
\begin{table}[]
\centering
\resizebox{\columnwidth}{!}{
\begin{tabular}{lllllc}
\toprule
\textbf{Dataset} & \textbf{Statistics} & \textbf{Train} & \textbf{Dev} & \textbf{Test} & \textbf{Level} \\
\midrule
\multirow{3}{*}{\textbf{QALB-2014}}
 & Number of sents. & $19,411$ & $1,017$ & $968$ & L1 \\
 & Number of words. & $1,021,165$  & $54,000$ & $51,000$ & L1 \\
 & Number of error. & $306,000$ & $16,000$ & $16,000$ & L1 \\
\midrule
\multirow{3}{*}{\textbf{QALB-2015}}
 & Number of sents. & $310$ & $154$ & $920$ & L2 \\
 & Number of words. & $43,353$ & $24,742$ & $48,547$ & L2 \\
 & Number of error. & $13,200$  & $7,300$ & $13,000$ & L2 \\
\bottomrule
\end{tabular}}
\caption{Statistics for  QALB-2014 and 2015 Train, development (Dev), and Test datasets.}
\label{tab:DatasetDescriptions}
\end{table}

%% file: Setups.tex

\section{LLMs and Prompting Techniques}\label{LLM}
This section outlines our experiments designed to instruction fine-tune LLMs and explore different prompting methods for ChatGPT in the context of AGEC. We begin by experimenting with various prompting strategies using ChatGPT, comparing its performance against smaller LLMs and our listed baselines. We evaluate the performance of ChatGPT-3.5 Turbo (ChatGPT) and GPT-4, under two prompting strategies: \textbf{\textit{Few-shot CoT}}~\cite{fang2023chatgpt} and \textbf{\textit{Expert Prompting}}~\cite{xu2023expertprompting}. We now describe our prompting strategies.


\noindent\subsection{ChatGPT Prompting}

\noindent\textbf{Preliminary experiment.} 
Initially, we experiment with a diverse set of prompt templates to assess ChatGPT’s capabilities in zero-shot learning as well as two aspects of few-shot learning: vanilla few-shot and few-shot CoT ~\cite{fang2023chatgpt}. We also experiment with prompts in both English and Arabic. However, we discover that the responses from these prompt templates contain extraneous explanations and are disorganized, necessitating substantial preprocessing for compatibility with the M$^2$ scorer. This problem is particularly notable in the zero-shot and Arabic prompt setups, which fails to yield output we can automatically evaluate. 

\noindent\textbf{Few-shot CoT.} 
Adopting the few-shot CoT prompt design strategy from~\citet{kojima2022large} and~\citet{fang2023chatgpt}, we implement a two-stage approach. Initially, we engage in \textit{`reasoning extraction'}, prompting the model to formulate an elaborate reasoning pathway. This is followed by an \textit{`answer extraction'} phase, where the reasoning text is combined with an answer-specific trigger sentence to form a comprehensive prompt. In our few-shot CoT settings, we include labeled instances from the Dev set in our prompts to implement ICL, facilitating learning from examples~\cite{brown2020language}. This involves providing erroneous sentences, labeled \colorbox{red!10}{\textit{<input> SRC </input>}}, along with their corrected versions, labeled \colorbox{green!10}{\textit{<output> TGT </output>}}, from the original Dev set.

\begin{figure*}[]
    \centering
    \includegraphics[width=0.9\textwidth]{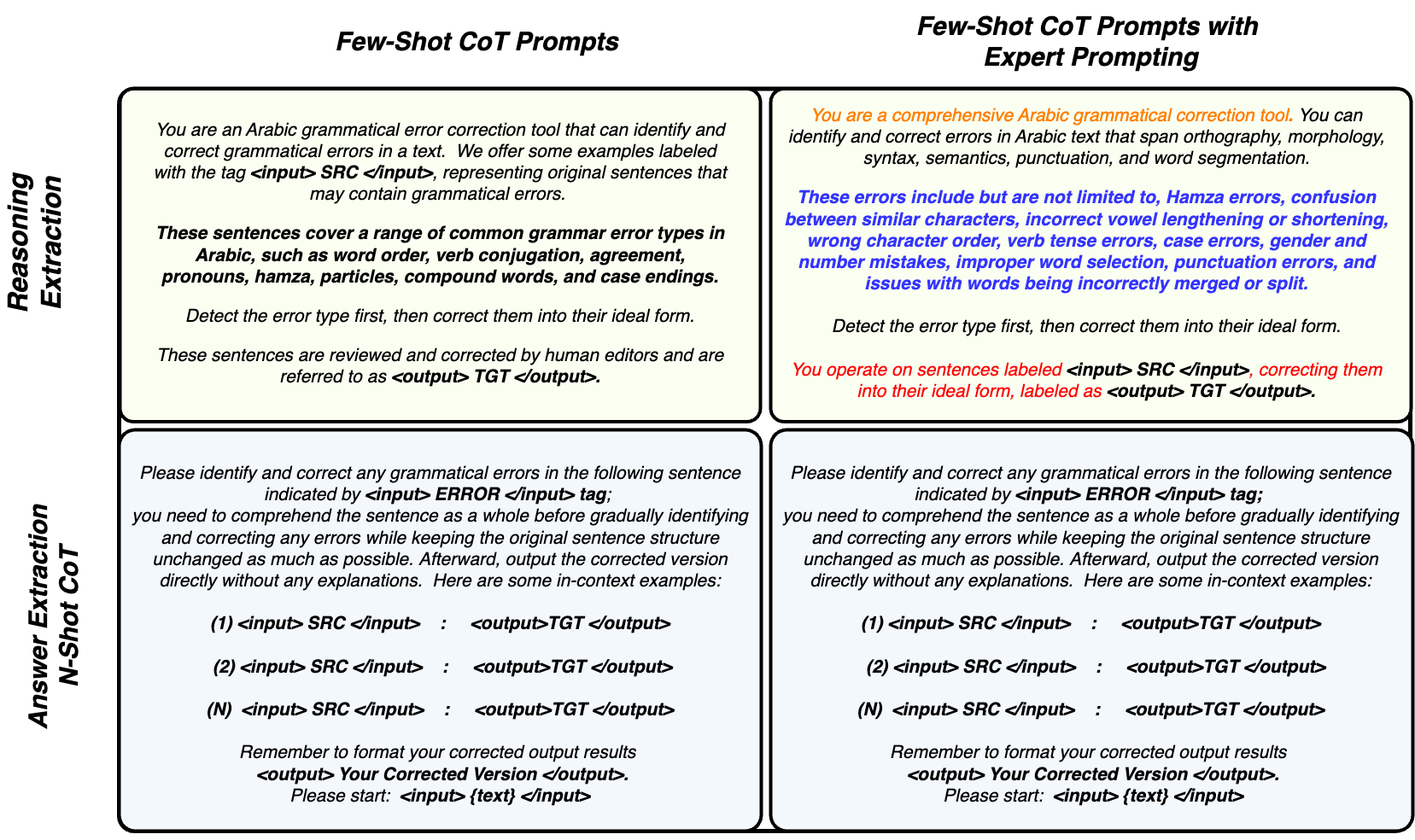}
    \caption{Illustration of Few-Shot CoT and Expert Prompts for Arabic Grammatical Error Correction.}
    \label{fig:x cubed graph}
\end{figure*}

\noindent\textbf{Expert prompting.}
\citet{xu2023expertprompting} introduces a novel strategy, which leverages the expert-like capabilities of LLMs. This method involves assigning expert personas to LLMs, providing specific instructions to enhance the relevance and quality of the generated responses. Following the framework of \citet{xu2023expertprompting}, we ensure that our AGEC correction tool exhibits three key characteristics: being \textcolor{orange}{\textit{\textbf{distinguished}}}, \textcolor{blue}{\textit{\textbf{informative}}}, and \textcolor{red}{\textit{\textbf{automatic}}} during the \textit{`reasoning extraction'} stage of our prompt. To achieve this, we employ a distinct and informative collection of various error types as proposed in the Arabic Learner Corpus taxonomy~\cite{alfaifi2012arabic}. We then prompt to automate the system by instructing it to operate on sentences labeled with \colorbox{red!10}{\textit{<input>}} and \colorbox{green!10}{\textit{<output>}} tags. Both prompts are illustrated in Figure \ref{fig:x cubed graph}.

\subsection{ChatGPT Results.} Table~\ref{tab:ChatGPT_eval} presents the performance of ChatGPT under different prompting strategies, compared to the baseline settings. We observe improvements, particularly as we progress from the one-shot to five-shot configurations for both the few-shot CoT and expert prompting (EP) strategies. Under the CoT prompt, ChatGPT's F\textsubscript{1.0} score increases from $53.59$ in the one-shot setting to $62.04$ in the five-shot setting. A similar upward trend is evident with the EP strategy, where the F\textsubscript{1.0} score rises from $55.56$ (one-shot) to $63.98$ (five-shot). Among all experiments involving ChatGPT, the three-shot and five-shot settings of GPT-4, CoT, achieve the highest scores, with F\textsubscript{1.0} of $63.98$ and $65.49$, respectively.

\input{Tables/ChatGPT_eval}

\noindent\subsection{Instruction-Finetuning LLMs}

\noindent\textbf{Fine-tuning LLMs.} 
To instruct fine-tune \textit{relatively} large models, \textit{henceforth} just LLMs, we first train these models on the translated Alpaca dataset~\cite{alpaca}~\footnote{We translate the Alpaca datasets using NLLB MT model~\cite{costa2022no}} to allow the models to gain deeper understanding of the Arabic language and its complexities. Following this, we further fine-tune the models on the QALB dataset, to specifically target the task of GEC. Then, we employ well-structured task instructions and input prompts, enabling the models to take on GEC tasks. Each model is assigned a task, given an instruction and an input for output generation. We provide an illustration of the instructions we use for model training in Appendix \ref{app:IF}.

\begin{figure}[t]
\centering
\includegraphics[width=1\columnwidth]{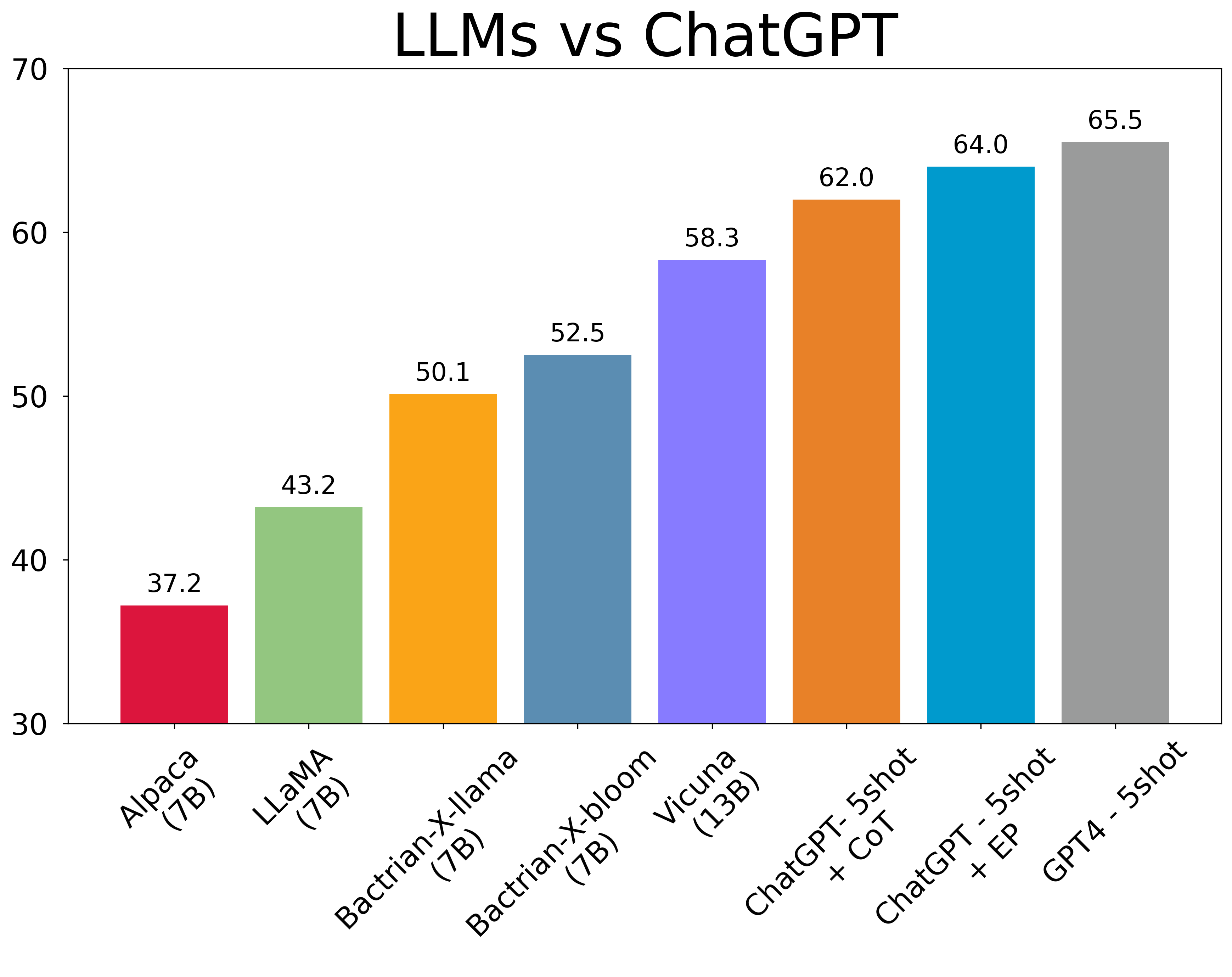}
\caption{Comparison of F\textsubscript{$1$} scores between LLMs and ChatGPT on the QALB-2014 Test set.}
\label{fig:gptllm}
\end{figure}

\noindent\textbf{LLM results.} 
As shown in Figure~\ref{fig:gptllm}, larger models such as Vicuna-13B and models trained on multilingual datasets like Bactrian-X$_{\textit{llama}}$-7B, and Bactrian-X$_{\textit{bloom}}$-7B exhibit an overall trend of better performance, achieving F\textsubscript{1} of $58.30$, $50.1$, and $52.5$, respectively. Despite these improvements, it is noteworthy that all models fall short of ChatGPT's. This reaffirms ChatGPT's superior ability on AGEC.

\section{Data Augmentation} \label{DA}
Motivated by the significant improvements observed in low-resource GEC tasks in languages such as German, Russian, and Czech through synthetic data~\cite{flachs-etal-2021-data}, and recognizing the recent efforts to develop synthetic data for AGEC~\cite{SOLYMAN2021303}, we experiment with three distinctive data augmentation methods. 

\noindent\textbf{ChatGPT as corruptor.}
With slight adaptation to our original  prompt, we engage ChatGPT as an AI model with the role of introducing grammatical errors into Arabic text to generate artificial data. We randomly sample 10,000 correct sentences from the QALB-2014 Train set and, using the taxonomy put forth by the Arabic Learner Corpus~\cite{alfaifi2012arabic}, prompt ChatGPT to corrupt these, creating a parallel dataset. We refer to the resulting dataset as {\bf syntheticGPT}. 



\noindent\textbf{Reverse noising.}
We adopt a \textit{reverse noising} approach~\cite{xie-etal-2018-noising}, training a reverse model that converts clean sentences \textit{Y} into noisy counterparts \textit{X}. This involves implementing a standard beam search to create noisy targets $\hat{Y}$ from clean input sentences \textit{Y}. Our approach incorporates two types of reverse models: the first trains on both QALB-2014 and 2015 gold datasets, and the second on the syntheticGPT dataset. Subsequently we generate a parallel dataset using commentaries from the same newspaper domain as our primary clean inputs, matching the original Train data. We name the respective parallel datasets {\bf reverseGold}, and {\bf reverseGPT}.


\begin{figure*}[!h]
\centering
\includegraphics[width=\textwidth]{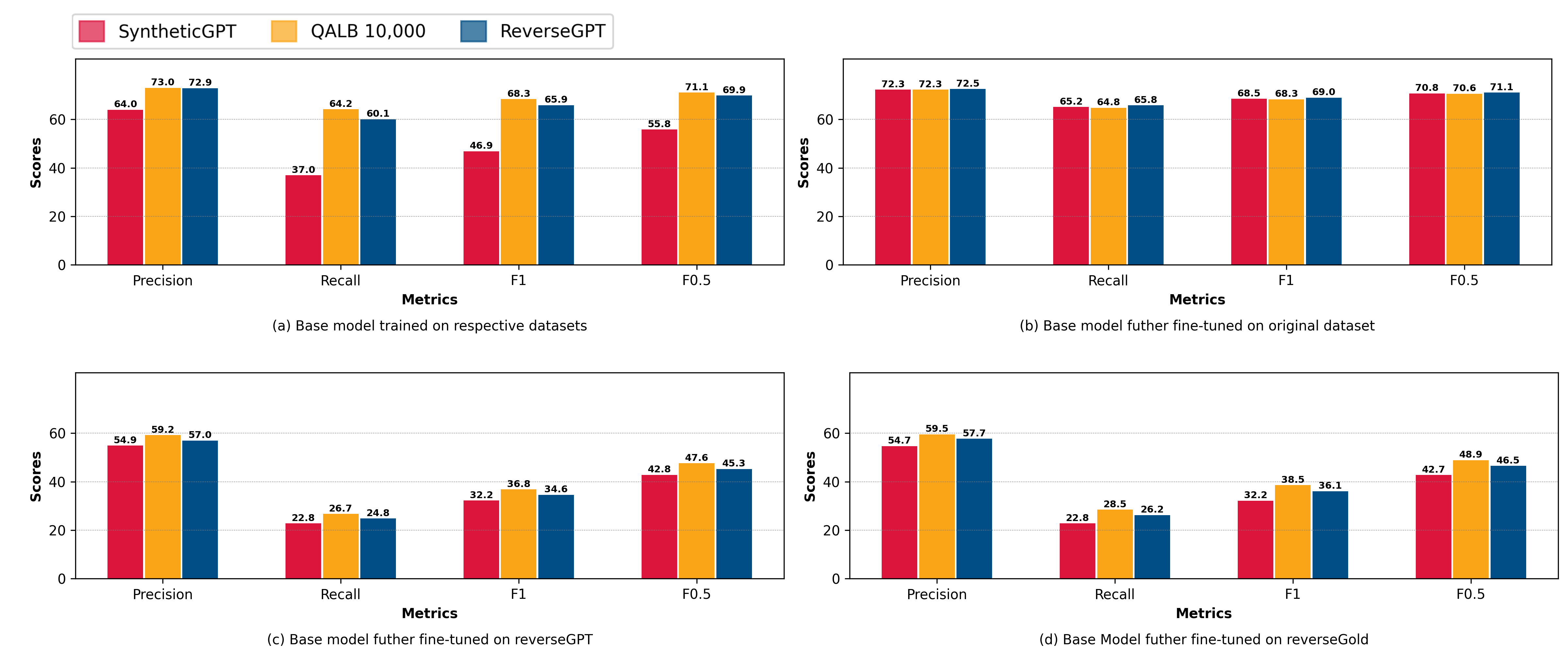}
\caption{Scores of models fine-tuned on $10,000$ parallel sentences from different sources: Original training data, syntheticGPT, and reverseGPT evaluated on the QALB-2014 Test set.}
\label{fig:three}
\end{figure*}

\noindent\textbf{Data augmentation evaluation.}
To evaluate the efficacy of ChatGPT in generating artificial data, we select $10,000$ parallel sentences from syntheticGPT, $10,000$ examples from reverseGPT, and $10,000$ parallel sentences from the original training set. We then further fine-tune each model on the original training dataset and the two synthetically generated reverse noised datasets, aiming to assess if these artificially crafted datasets can replace the gold standard training set. Figure~\ref{fig:three} shows our results. In our initial tests (Figure~\ref{fig:three}.a), fine-tuning the AraT5\textsubscript{v2} model exclusively on the $10,000$ sentences from syntheticGPT, registers an F\textsubscript{1} of $65.87$, and reverseGPT an F\textsubscript{1} score of $46.85$ falling behind the original QALB 2014 training data (which records an F\textsubscript{1} of $68.34$). Following this, when further fine-tuned on the original training set (Figure~\ref{fig:three}.b). We find that both syntheticGPT and the reverseGPT surpass model fine-tuned on equivalent-sized gold dataset, with F\textsubscript{1} of $69.01$ and $68.54$, respectively. This confirms the utility of ChatGPT for generating synthetic data. Conversely, when we further fine-tune the model with the two reverse noised datasets  (see Figures~\ref{fig:three}.c and d), we observe a sharp decline in performance.
This emphasizes the critical importance of relevant, high-quality synthetic data over randomly generated samples.

\subsection{Decoding Methods.}
Decoding strategies for text generation are essential and can vary based on the task~\cite{zhang2023llamaadapter}. We compare three decoding strategies to identify the best method for AGEC task. As shown in Table~\ref{tab:AraT5_eval_new}, we compare \textit{greedy decoding}~\cite{germann-2003-greedy} (temperature=0), \textit{Beam search}~\cite{freitag2017beam} (num\_beams=5, temperature=1), and \textit{Top-P sampling}~\cite{holtzman2019curious} (top-p=0.8, top-k=75, and temperature=0.8). With the highest scoring strategy identified, we scale up our data augmentation experiments, by generating sets of 5million and 10million reverseGold datasets. In addition to these datasets, we utilize the complete AGEC dataset from~\citet{SOLYMAN2021303} (referred to as AraT5\textsubscript{v2} (11M) in our experiments) for further evaluation.

Outlined in Table~\ref{tab:AraT5_eval_sorted}, AraT5\textsubscript{v2} shows consistent 
improvement as the number of training samples increases from 5M to 11M. Results indicate Top-P sampling is the best decoding method for GEC, exhibiting a balance between number of correct edits and total number of edits made.

\input{Tables/AraT5_eval_new}
\input{Tables/sd_mean}

\section{Sequence Tagging Approach} \label{ST}
In this section, we detail our methods to adapt the GECToR model~\cite{omelianchuk2020gector} to experiment with the seq2edit approach.

\noindent\textbf{Token-level transformations.}
We first perform token-level transformations on the source to recover the target text.~\textit{\textbf{`Basic-transformations'}} are applied to perform the most common token-level edit operations, such as keeping the current token unchanged (\texttt{\$KEEP}), deleting current token (\texttt{\$DELETE}), appending new token  t\_\textsubscript{$1$} next to the current token x\textsubscript{i} (\texttt{\$APPEND\_\texttt{t}\textsubscript{$1$}}) or replacing the current token x\textsubscript{i} with another token t\_\textsubscript{$2$} (\texttt{\$REPLACE\_\texttt{t}\textsubscript{$2$}}). To apply tokens with more task-specific operations, we employ \textit{\textbf{`g-transformations'}} such as  the (\texttt{\$MERGE}) tag to merge the current token and the next token into a single one. Edit space after applying token-level transformations results in \texttt{KEEP} (\texttt{$725$K op}),  
 \texttt{\$REPLACE\_\texttt{t}\textsubscript{$2$}} (\texttt{$201$K op}), \texttt{\$APPEND\_\texttt{t}\textsubscript{$1$}} (\texttt{$75$K op}), \texttt{\$DELETE} (\texttt{$13$K op}), and \texttt{\$MERGE} (\texttt{$5.7$K op}) tags.  



\noindent\textbf{Preprocessing and fine-tuning.}
We start the preprocessing stage by aligning source tokens with target subsequences, preparing them for token-level transformations. We then fine-tune ARBERT\textsubscript{v2} \cite{elmadany2022orca} and MARBERT\textsubscript{v2}~\cite{abdul-mageed-etal-2021-arbert} on the preprocessed data. We adhere to the training approach detailed in the original paper~\cite{omelianchuk2020gector}, adopting its three-stage training and setting the iterative correction to three. More details about the fine-tuning procedure can be found in Appendix~\ref{app:GECToR}.


\noindent\textbf{Sequence tagging evaluation.}
As shown in Table~\ref{tab: Gector_Results}, ARBERT\textsubscript{v2} and MARBERT\textsubscript{v2}, exhibit high precision (e.g., ARBERT\textsubscript{v2}'s three-step training is at $74.39$ precision). However, relatively lower recall scores indicate challenges in ability of the two models to detect errors. Unlike the findings in the original paper, our implementation of a three-stage training approach yields mixed results: while accuracy improves, recall scores decrease, leading to a drop in the overall F\textsubscript{1} score (by $0.36$ for ARBERT\textsubscript{v2} and $1.10$ for MARBERT\textsubscript{v2}, respectively). Consequently, all models fall behind the 'seq2seq' models. We note that both ARBERT\textsubscript{v2} and MARBERT\textsubscript{v2} surpass mT0 and mT5 in terms of F\textsubscript{0.5} scores, highlighting their abilities in correcting errors with precision.

\input{Tables/GECTor_Results}

\input{Tables/Errors_Types_Examples}

\section{Error Analysis} \label{AET} 
\subsection{Error type evaluation.} 
\begin{figure*}[h]
\centering
\includegraphics[width=0.95\textwidth]{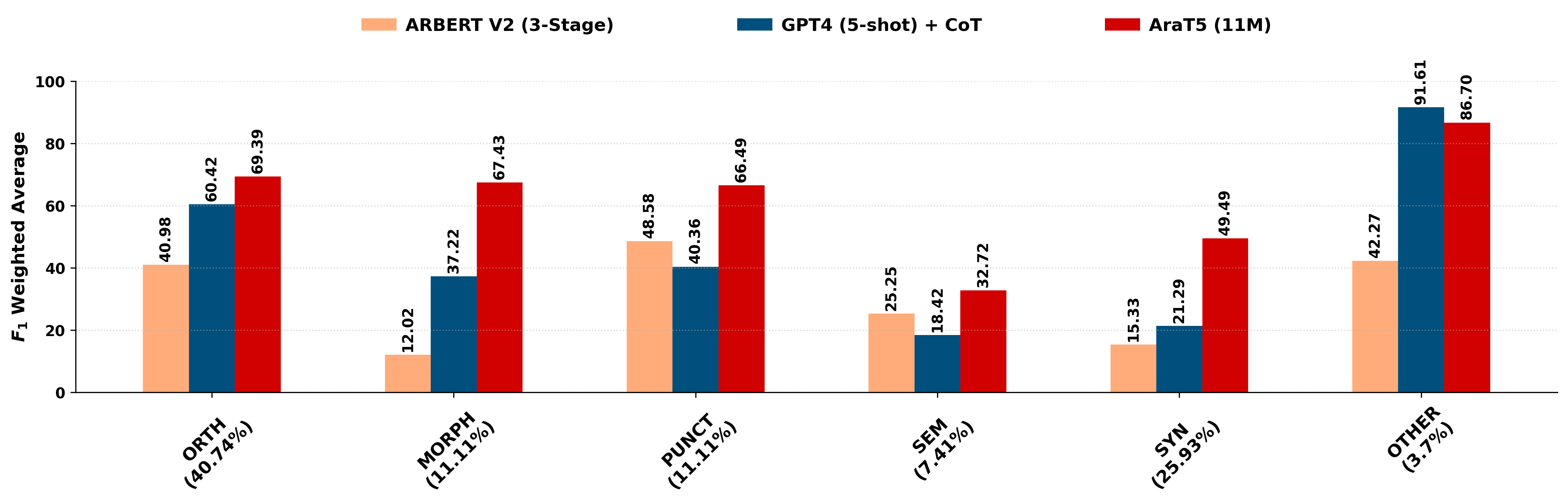}
\caption{Best model F\textsubscript{1} scores for each approach on specific error types in the QALB-2014 Test set.}
\label{fig:areta_plot}
\end{figure*}

We use the Automatic Error Type Annotation (ARETA) tool~\cite{belkebir2021automatic} to assess our models' performance on different error types. We focus on seven errors types: \textit{Orthographic}, \textit{Morphological}, \textit{Syntactic}, \textit{Semantic}, \textit{Punctuation}, \textit{Merge}, and \textit{Split}. Examples of each error types alongside their translations can be found in Table~\ref{tab:errors_type_examples}. We examine top models from each approach, including ARBERT\textsubscript{v2} (3-step), GPT-4 (5-shot) + CoT, and AraT5\textsubscript{v2}(11M). Figure~\ref{fig:areta_plot} illustrates the performance of selected models under each error type. AraT5\textsubscript{v2}(11M), surpasses all other models across all error categories. In particular, it excels in handling \textit{Orthographic} (\text{ORTH}) errors, \textit{Morphological} (\text{MORPH}) errors, and \textit{Punctuation} (\text{PUNCT}) errors, consistently achieving over $65$ F\textsubscript{1} score. However, it is worth observing that all models encounter challenges with \textit{Semantic} (\text{SEM}) and \textit{Syntactic} (\text{SYN}) errors. These disparate outcomes underscore the significance of selecting the appropriate model based on the error types prevalent in a specific dataset.

\subsection{Normalization methods.}
\noindent In addition to the \textit{`Exact Match'} score, we also analyze system performance under different normalization methods. Namely, we assess the system on normalized text (1) without Alif/Ya errors, (2) without punctuation, and (3) free from both  Alif/Ya and punctuation errors. Examples of text under each setting  can be found in Appendix~\ref{app:Error_type_examples}.

\subsection{Normalisation results}
\input{Tables/2014_2015_v2}
Looking at Table~\ref{tab: v2}, in the `No punctuation' setting, all models perform better, reflecting models' limitations in handling punctuation which is due to absence of clearly agreed upon punctuation rules in Arabic.  Moreover, the datasets used are based on commentaries where punctuation is inherently inconsistent and varied. Another noteworthy observation is the drop in F\textsubscript{1} scores when Alif/Ya errors are removed. This can be attributed to the fact that Alif/Ya errors are relatively simpler compared to other error categories. Moreover, AraT5\textsubscript{v2} is trained on formal texts such as AraNews~\cite{nagoudi2020machine} and Hindawi Books~\footnote{www.hindawi.org/books}, which contain proper Alif/Ya indicating the model's proficiency with the correct usage of these letters.

%% file: Tables/ChatGPT_eval.tex
\begin{table}[t]
\centering
\resizebox{\columnwidth}{!}{%
\begin{tabular}{llcccc}\toprule
\multirow{2}{*}{\textbf{Settings}} &\multirow{2}{*}{\textbf{Models}} &\multicolumn{4}{c}{\textbf{Exact Match}} \\\cmidrule{3-6}
& &\textbf{P} &\textbf{R} &\textbf{F\textsubscript{1.0}} &\textbf{F\textsubscript{0.5}} \\ \toprule 
\multirow{5}{*}{\textbf{Baselines}} 
&mT0 & $70.76$~\textsuperscript{$\pm$0.03} & $50.78$~\textsuperscript{$\pm$0.07} & $59.12$~\textsuperscript{$\pm$0.05} & $65.59$~\textsuperscript{$\pm$0.03} \\    
&mT5 &$70.64$~\textsuperscript{$\pm$0.12} &$50.16$~\textsuperscript{$\pm$0.05} &$58.66$~\textsuperscript{$\pm$0.05} &$65.30$~\textsuperscript{$\pm$0.09} \\
&AraBART &$70.71$~\textsuperscript{$\pm$0.06} &$60.46$~\textsuperscript{$\pm$0.04} &$65.18$~\textsuperscript{$\pm$0.07} &$68.39$~\textsuperscript{$\pm$0.08} \\
&AraT5\textsubscript{v2} &$\mathbf{73.04}$~\textsuperscript{$\pm$0.10} &$\mathbf{63.09}$~\textsuperscript{$\pm$0.15} &$\mathbf{67.70}$~\textsuperscript{$\pm$0.12} &$\mathbf{70.81}$~\textsuperscript{$\pm$0.11} \\
\midrule
\multirow{3}{*}{\textbf{+ CoT}} 
&ChatGPT (1-shot) &$58.71$ &$49.29$ &$53.59$ &$56.55$ \\
&ChatGPT (3-shot) &$64.60$ &$60.37$ &$62.41$ &$63.71$ \\
&ChatGPT (5-shot) &$64.70$ &$59.59$ &$62.04$ &$63.61$ \\
\midrule
\multirow{3}{*}{\textbf{+ EP}} 
&ChatGPT (1-shot) &$60.49$ &$51.37$ &$55.56$ &$58.42$ \\
&ChatGPT (3-shot) &$65.83$ &$61.41$ &$63.54$ &$64.90$ \\
&ChatGPT (5-shot) &$66.53$ &$61.62$ &$63.98$ &$65.49$ \\
\midrule
\multirow{3}{*}{\textbf{+ CoT}} 
&GPT4 (1-shot) \textsuperscript{$*$} & $-$  & $-$  & $-$  & $-$  \\
&GPT4 (3-shot) &$69.31$ &$59.24$ &$63.88$ &$67.03$ \\
&GPT4 (5-shot) &$69.46$ &$61.96$ &$65.49$ &$67.82$ \\
\toprule
\end{tabular}%
}
\caption{Performance of ChatGPT under different prompting strategies on QALB-2014 Test set.\textsuperscript{$^*$}Results for QALB-2015 Test and GPT4 1-shot are not included due to the high cost in producing these results, and a pattern has already been established showing that performance increases as we increase the number of N-shot examples. More details are in Appendix~\ref{app:BES}.} \label{tab:ChatGPT_eval}
\end{table}


%% file: Tables/AraT5_eval_new.tex
\begin{table}[t]
\centering
\renewcommand{\arraystretch}{1.5}
\resizebox{\columnwidth}{!}{%
\begin{tabular}{llcccc|cccc}
\toprule
\multirow{2}{*} & \multirow{2}{*}{\textbf{Strategy}} & \multicolumn{4}{c}{\textbf{QALB-2014}} & \multicolumn{4}{c}{\textbf{QALB-2015}} \\
\cmidrule(lr){3-6} \cmidrule(lr){7-10}
& & \textbf{P} & \textbf{R} & \textbf{F\textsubscript{1}} & \textbf{F\textsubscript{0.5}} & \textbf{P} & \textbf{R} & \textbf{F\textsubscript{1}} & \textbf{F\textsubscript{0.5}} \\
\midrule
\multirow{3}{*}
& \textbf{Greedy} & $74.09\textsuperscript{$\pm$0.57}$ & $65.63\textsuperscript{$\pm$0.59}$ & $69.60\textsuperscript{$\pm$0.54}$ & $72.23\textsuperscript{$\pm$0.55}$ & $67.41\textsuperscript{$\pm$0.82}$ &$ 66.85\textsuperscript{$\pm$0.97}$ & $67.13\textsuperscript{$\pm$0.82}$ & $67.30\textsuperscript{$\pm$0.80}$ \\
& \textbf{Beam} & $75.47\textsuperscript{$\pm$1.11}$ & $68.61\textsuperscript{$\pm$1.26}$ & $71.87\textsuperscript{$\pm$1.19}$ &$ 73.99\textsuperscript{$\pm$1.14}$ & $70.54\textsuperscript{$\pm$0.44}$ & $68.04\textsuperscript{$\pm$0.14}$ & $69.27\textsuperscript{$\pm$0.24}$ & $70.03\textsuperscript{$\pm$0.35}$ \\
& \textbf{Top-p} & $76.94\textsuperscript{$\pm$0.67}$ & $69.26\textsuperscript{$\pm$0.73}$ & $72.90\textsuperscript{$\pm$0.68}$ & $75.27\textsuperscript{$\pm$0.67}$ & $72.64\textsuperscript{$\pm$0.32}$ & $74.21\textsuperscript{$\pm$0.75}$ & $73.41\textsuperscript{$\pm$0.51}$ & $72.94\textsuperscript{$\pm$0.39}$ \\
\bottomrule
\end{tabular}%
 }
\caption{Performance of AraT5\textsubscript{v2} (11M) on QALB-2014 and 2015 Test set under different decoding methods.}
\label{tab:AraT5_eval_new}
\end{table}

%% file: Tables/sd_mean.tex
\begin{table}[t]
\centering
\renewcommand{\arraystretch}{1.5}
\resizebox{\columnwidth}{!}{%
\begin{tabular}{lcccc|cccc}
\toprule
\multirow{2}{*}{\textbf{Datasets}}  & \multicolumn{4}{c}{\textbf{QALB-2014}} & \multicolumn{4}{c}{\textbf{QALB-2015}} \\
\cmidrule(lr){2-5} \cmidrule(lr){6-9}
&  \textbf{P} & \textbf{R} & \textbf{F\textsubscript{1}} & \textbf{F\textsubscript{0.5}} & \textbf{P} & \textbf{R} & \textbf{F\textsubscript{1}} & \textbf{F\textsubscript{0.5}} \\
\midrule

\textbf{\colorbox{green!20}{M1}} & $71.35$\textsuperscript{$\pm$0.14} & $64.45$\textsuperscript{$\pm$0.41} & $67.73$\textsuperscript{$\pm$0.17} & $69.85$\textsuperscript{$\pm$0.04} & $69.65$\textsuperscript{$\pm$0.57} & $64.74$\textsuperscript{$\pm$0.57} & $67.11$\textsuperscript{$\pm$0.14} & $68.61$\textsuperscript{$\pm$0.33} \\
\textbf{\colorbox{red!10}{M2}} & $73.14$\textsuperscript{$\pm$0.26} & $67.48$\textsuperscript{$\pm$1.07} & $70.23$\textsuperscript{$\pm$0.15} & $72.37$\textsuperscript{$\pm$1.05} & $70.26$\textsuperscript{$\pm$1.16} & $65.74$\textsuperscript{$\pm$1.37} & $67.93$\textsuperscript{$\pm$1.27} & $69.31$\textsuperscript{$\pm$1.20} \\
\textbf{\colorbox{yellow!25}{M3}} & $\textbf{76.94}\textsuperscript{$\pm$0.67}$ & $\textbf{69.26}\textsuperscript{$\pm$0.73}$ & $\textbf{72.90}\textsuperscript{$\pm$0.68}$ & $\textbf{75.27}\textsuperscript{$\pm$0.67}$ & $\textbf{72.64}\textsuperscript{$\pm$0.32}$ & $\textbf{74.21}\textsuperscript{$\pm$0.75}$ & $\textbf{73.41}\textsuperscript{$\pm$0.51}$ & $\textbf{72.94}\textsuperscript{$\pm$0.39}$ \\
\bottomrule
\end{tabular}%
}
\caption{Performance of AraT5\textsubscript{v2} models using the 'Top-P' decoding method on QALB-2014 and 2015 Test sets, on different amounts of training data. \colorbox{green!20}{M1}: AraT5\textsubscript{v2} (5M), \colorbox{red!10}{M2}: AraT5\textsubscript{v2} (10M), \colorbox{yellow!25}{M3}: AraT5\textsubscript{v2} (11M)}
\label{tab:AraT5_eval_sorted}
\end{table}

%% file: Tables/GECTor_Results.tex
\begin{table}[]
\centering
\renewcommand{\arraystretch}{1.5}
\resizebox{\columnwidth}{!}{%
\begin{tabular}{llcccc|cccc}\toprule
\multirow{2}{*}{\textbf{Methods}} &\multirow{2}{*}{\textbf{Models}} &\multicolumn{4}{c}{\textbf{QALB-2014}} &\multicolumn{4}{c}{\textbf{QALB-2015}}\\\cmidrule{3-6}\cmidrule{7-10}
& &\textbf{P} &\textbf{R} &\textbf{F\textsubscript{1.0}} &\textbf{F\textsubscript{0.5}} &\textbf{P} &\textbf{R} &\textbf{F\textsubscript{1.0}} &\textbf{F\textsubscript{0.5}}\\\toprule
\multirow{4}{*}{\textbf{Seq2Seq}} 
&mT0 & $70.76$~\textsuperscript{$\pm$0.03} & $50.78$~\textsuperscript{$\pm$0.07} & $59.12$~\textsuperscript{$\pm$0.05} & $65.59$~\textsuperscript{$\pm$0.03}&$68.11$~\textsuperscript{$\pm$0.20} &$59.68$~\textsuperscript{$\pm$0.12} &$63.61$~\textsuperscript{$\pm$0.15} &$66.23$~\textsuperscript{$\pm$0.18}  \\    
&mT5 &$70.64$~\textsuperscript{$\pm$0.12} &$50.16$~\textsuperscript{$\pm$0.05} &$58.66$~\textsuperscript{$\pm$0.05} &$65.30$~\textsuperscript{$\pm$0.09}  &$68.20$~\textsuperscript{$\pm$0.10} &$59.02$~\textsuperscript{$\pm$0.15} &$63.28$~\textsuperscript{$\pm$0.04} &$66.14$~\textsuperscript{$\pm$0.11} \\
&AraBART &$70.71$~\textsuperscript{$\pm$0.06} &$60.46$~\textsuperscript{$\pm$0.04} &$65.18$~\textsuperscript{$\pm$0.07} &$68.39$~\textsuperscript{$\pm$0.08} &$68.39$~\textsuperscript{$\pm$0.09} &$67.95$~\textsuperscript{$\pm$0.02} &$65.62$~\textsuperscript{$\pm$0.05} &$66.76$~\textsuperscript{$\pm$0.07} \\
&AraT5\textsubscript{v2} &${73.04}$~\textsuperscript{$\pm$0.10} &$\mathbf{63.09}$~\textsuperscript{$\pm$0.15} &$\mathbf{67.70}$~\textsuperscript{$\pm$0.12} &$\mathbf{70.81}$~\textsuperscript{$\pm$0.11} &$71.40$~\textsuperscript{$\pm$0.90} &$\mathbf{72.83}$~\textsuperscript{$\pm$1.11} &$\mathbf{72.11}$~\textsuperscript{$\pm$0.99} &$\mathbf{71.68}$~\textsuperscript{$\pm$0.93}  \\\midrule
\multirow{4}{*}{\textbf{Seq2edit}} 
&ARBERTv2 &${73.89}$~\textsuperscript{$\pm$0.35} &${48.33}$~\textsuperscript{$\pm$0.33} &${58.43}$~\textsuperscript{$\pm$0.35} &${66.82}$~\textsuperscript{$\pm$0.35} &${73.10}$~\textsuperscript{$\pm$0.29} &${55.40}$~\textsuperscript{$\pm$1.15} &${63.03}$~\textsuperscript{$\pm$0.86} &${68.70}$~\textsuperscript{$\pm$0.56} \\
&{ARBERT\textsubscript{v2}}\textsuperscript{$\dagger$}
&$\mathbf{74.39}$~\textsuperscript{$\pm$0.22} &${47.62}$~\textsuperscript{$\pm$0.30} &${58.07}$~\textsuperscript{$\pm$0.29} &${66.87}$~\textsuperscript{$\pm$0.26} &$\mathbf{74.20}$~\textsuperscript{$\pm$0.28} &${53.80}$~\textsuperscript{$\pm$0.59} &${62.37}$~\textsuperscript{$\pm$0.49} &${68.96}$~\textsuperscript{$\pm$0.39} \\
&MARBERT\textsubscript{v2} &${73.53}$~\textsuperscript{$\pm$0.24} &${48.21}$~\textsuperscript{$\pm$0.39} &${58.24}$~\textsuperscript{$\pm$0.36} &${66.54}$~\textsuperscript{$\pm$0.30} &${72.90}$~\textsuperscript{$\pm$0.21} &${54.90}$~\textsuperscript{$\pm$0.52} &${62.63}$~\textsuperscript{$\pm$0.42} &${68.41}$~\textsuperscript{$\pm$0.31} \\
&{MARBERT\textsubscript{v2}}\textsuperscript{$\dagger$}
&${74.21}$~\textsuperscript{$\pm$0.16} &${46.45}$~\textsuperscript{$\pm$0.25} &${57.14}$~\textsuperscript{$\pm$0.24} &${66.29}$~\textsuperscript{$\pm$0.20} &${74.00}$~\textsuperscript{$\pm$0.17} &${52.70}$~\textsuperscript{$\pm$0.34} &${61.56}$~\textsuperscript{$\pm$0.29} &${68.46}$~\textsuperscript{$\pm$0.23} \\
\bottomrule
\end{tabular}%
}
\caption{Performance of the seq2edit approach compared to baselines on the QALB-2014 and QALB-2015 Test sets. \textsuperscript{$\dagger$}: Models trained on 3-stage training.}\label{tab: Gector_Results}
\end{table}

%% file: Tables/Errors_Types_Examples.tex
\begin{table}[t]
\centering
\renewcommand{\arraystretch}{1.5}
 \resizebox{\columnwidth}{!}{%
\begin{tabular}{lHrr}
\toprule
\textbf{Error Type} & Description& \multicolumn{1}{c}{\textbf{Incorrect Sentence}} & \multicolumn{1}{c}{\textbf{Correct Sentence}}   \\ \toprule
\multirow{2}{*}{\textbf{Orthographic} }  & \multirow{2}{*}{}                       
&  \< الفرس . > \textcolor{red}{ \< يرب > }    \< الرجل >  
&   \< الفرس . > \textcolor{green}{ \< يركب > }    \< الرجل >   
\\
&                                         & \textit{The man \textcolor{red}{rears} the horse.}                         & \textit{The man \textcolor{green}{rides} the horse. }  \\ 
\multirow{2}{*}{\textbf{Punctuations}}  & \multirow{2}{*}{}                       
&  \<  يركب الفرس . >  \textcolor{red}{ \< ، > }    \< الرجل >  
& \<الرجل يركب الفرس .>  
    \\
                                      &                                          & The man\textcolor{red}{,} rides the horse.                      & The man rides the horse.                        \\

\multirow{2}{*}{\textbf{Syntax}}        & \multirow{2}{*}{}      
&  \textcolor{red}{ \< فرس .>  } \<وجد رجلا يركب>   
&  \textcolor{green}{ \< فرسا .>  } \<وجد رجلا يركب>   

\\
  &                                          & He found a man riding a \textcolor{red}{hors}.                  & He found a man riding a \textcolor{green}{horse}.                  \\
                                      
\multirow{2}{*}{\textbf{Merge}}         & \multirow{2}{*}{}                    

&  \< الفرس . >  \< سيركب > \textcolor{red}{ \< غداالرجل >  }
&  \< الفرس . >  \< سيركب > \textcolor{green}{ \< غدا الرجل >  }
\\

                                      &                                          & \textcolor{red}{Tomorrowtheman} will ride the horse.             & \textcolor{green}{Tomorrow the man} will ride the horse.           \\


\multirow{2}{*}{\textbf{Splits}}        & \multirow{2}{*}{}                     
&  \<  الفرس . >   \textcolor{red}{ \< ير كب > }    \< غدا الرجل >  
&  \<  الفرس . >   \textcolor{green}{ \< يركب > }    \< غدا الرجل >  

\\                                      &                                          & The man \textcolor{red}{ri des} the horse.                       & The man \textcolor{green}{rides} the horse.                        \\

\multirow{2}{*}{\textbf{Semantic}}      &                                         
&  \< ظهر الفرس . >   \textcolor{red}{ \< في  > }    \<  الرجل يجلس>  
&  \< ظهر الفرس . >   \textcolor{green}{ \< على  > }    \<  الرجل يجلس>  
\\
                                      &                                          & The man is sitting \textcolor{red}{in} the horse's back.         & The man is sitting \textcolor{green}{on} the horse's back.         \\


\multirow{2}{*}{\textbf{Morphological}} &                                  
&  \<  الفرس . >   \textcolor{red}{ \< ركب > }    \< غدا الرجل >  
&  \<  الفرس . >   \textcolor{green}{ \< سيركب > }    \< غدا الرجل >  

\\
                                      &                                          & Tomorrow the man \textcolor{red}{rode} the horse.                & Tomorrow the man \textcolor{green}{will ride} the horse.          \\

                                      \toprule
\end{tabular}}
\caption{Examples of Merge, Morphological, Orthographic, Punctuation, Semantic, Split, and Syntactic errors, along with their corresponding corrections and English translations.}\label{tab:errors_type_examples}
\end{table}

%% file: Tables/2014_2015_v2.tex
\begin{table*}[!htp]
\centering
\scriptsize
\renewcommand{\arraystretch}{1.6}
\resizebox{\textwidth}{!}{
\begin{tabular}{llrrrr|rrrr|rrrr|rrrrr}\toprule
\multirow{2}{*}{\textbf{Test Set}} &\multirow{2}{*}{\textbf{~~~~~~~~~~~~~~Models}} &\multicolumn{4}{c}{\textbf{Exact Match}} &\multicolumn{4}{c}{\textbf{No Alif / Ya Errors}} &\multicolumn{4}{c}{\textbf{No Punctuation}} &\multicolumn{4}{c}{\textbf{No Punctuation and Alif / Ya Errors}} \\\cmidrule{3-18}
& &\textbf{P} &\textbf{R} &\textbf{F\textsubscript{1.0}} &\textbf{F\textsubscript{0.5}} &\textbf{P} &\textbf{R} &\textbf{F\textsubscript{1.0}} &\textbf{F\textsubscript{0.5}} &\textbf{P} &\textbf{R} &\textbf{F\textsubscript{1.0}} &\textbf{F\textsubscript{0.5}} &\textbf{P} &\textbf{R} &\textbf{F\textsubscript{1.0}} &\textbf{F\textsubscript{0.5}} \\\midrule
\multirow{6}{*}{\textbf{QALB-2014}}
&\citet{SOLYMAN2021303} &$\mathbf{79.06}$ &$65.79$ &$71.82$ &$\mathbf{75.99}$ &- &- &- &- &- &- &- &- &- &- &- &- \\
& \citet{mohit2014first} &$73.34$ &$63.23$ &$67.91$ &$71.07$ &$\mathbf{64.05}$ &$50.86$ &$56.7$ &$\mathbf{60.89}$ &$76.99$ &$49.91$ &$60.56$ &$69.45$ &$76.99$ &$49.91$ &$60.56$ &$69.45$ \\ 
&GPT4 (5-shot) &$69.46$ &$61.96$ &$65.49$ &$67.82$ &$58.44$ &$51.47$ &$54.73$ &$56.90$ &$74.59$ &$78.15$ &$76.33$ &$75.28$ &$60.06$ &$65.75$ &$62.78$ &$61.12$ \\
&ARBERT\textsubscript{v2} (3-step) & $74.17^{\pm 0.22}$ & $47.34^{\pm 0.30}$ & $57.79^{\pm 0.29}$ & $66.62^{\pm 0.26}$ & $64.90^{\pm 0.57}$ & $41.86^{\pm 0.24}$ & $50.89^{\pm 0.17}$ & $58.46^{\pm 0.33}$ & $76.90^{\pm 0.85}$s & $46.33^{\pm 0.58}$ & $57.83^{\pm 0.66}$ & $67.94^{\pm 0.75}$ & $56.66^{\pm 0.57}$ & $29.30^{\pm 0.61}$ & $38.62^{\pm 0.39}$ & $47.74^{\pm 0.03}$ \\
& AraT5\textsubscript{v2} (11m) & $76.94\textsuperscript{$\pm$0.67}$ & $\mathbf{69.26}\textsuperscript{$\pm$0.73}$ & $\mathbf{72.90}\textsuperscript{$\pm$0.68}$ & $75.27\textsuperscript{$\pm$0.67}$  & $62.42^{\pm 0.68}$ & $\mathbf{52.56}^{\pm 0.51}$ & $\mathbf{57.06}^{\pm 0.08}$ & $60.16^{\pm 0.38}$ & $\mathbf{86.52}^{\pm 0.50}$ & $\mathbf{82.90}^{\pm 0.17}$ & $\mathbf{84.67}^{\pm 0.25}$ & $\mathbf{85.77}^{\pm 0.39}$ & $\mathbf{79.44}^{\pm 0.51}$ & $\mathbf{67.40}^{\pm 0.53}$ & $\mathbf{72.92}^{\pm 0.52}$ & $\mathbf{76.70}^{\pm 0.52}$ \\
\cmidrule{2-18}
\multirow{6}{*}{\textbf{QALB-2015}}
&\citet{SOLYMAN2021303} &$\mathbf{80.23}$ &$63.59$ &$70.91$ &$\mathbf{76.24}$ &- &- &- &- &- &- &- &- &- &- &- &- \\
&\citet{rozovskaya2015second}  &$88.85$ &$61.76$ &$72.87$ &$81.68$ &$\mathbf{84.25}$ &$43.29$ &$\mathbf{57.19}$ &$\mathbf{70.84}$ &$\mathbf{85.8}$ &$\mathbf{77.98}$ &$\mathbf{81.7}$ &$\mathbf{84.11}$ &$\mathbf{80.12}$ &$\mathbf{58.24}$ &$\mathbf{67.45}$ &$\mathbf{74.52}$ \\
&ChatGPT (3-shot) + EP &$52.33$ &$47.57$ &$49.83$ &$54.10$ &$37.93$ &$39.97$ &$38.92$ &$32.95$ &$53.38$ &$56.63$ &$54.96$ &$54.00$ &$33.33$ &$46.77$ &$38.92$ &$35.36$ \\
&ARBERT\textsubscript{v2} (3-step) & $73.92^{\pm 0.28}$ & $53.15^{\pm 0.59}$ & $61.84^{\pm 0.49}$ & $68.56^{\pm 0.39}$ & $57.14^{\pm 0.21}$ & $39.17^{\pm 0.76}$ & $46.47^{\pm 0.47}$ & $52.34^{\pm 0.13}$ & $66.90^{\pm 0.17}$ & $61.50^{\pm 0.50}$ & $64.09^{\pm 0.28}$ & $65.74^{\pm 0.18}$ & $71.18^{\pm 0.16}$ & $39.00^{\pm 0.87}$ & $50.39^{\pm 0.75}$ & $61.09^{\pm 0.49}$ \\
&AraT5\textsubscript{v2} (11m) & $72.10^{\pm 0.31}$ & $\mathbf{73.59}^{\pm 0.70}$ & $\mathbf{72.84}^{\pm 0.40}$ & $72.40^{\pm 0.30}$ & $55.80^{\pm 0.30}$ & $\mathbf{43.51}^{\pm 0.50}$ & $48.89^{\pm 0.22}$ & $52.81^{\pm 0.11}$ & $85.82^{\pm 0.31}$ & $72.85^{\pm 0.25}$ & $78.81^{\pm 0.28}$ & $82.87^{\pm 0.30}$ & $75.08^{\pm 0.13}$ & $53.30^{\pm 0.93}$ & $62.34^{\pm 0.60}$ & $69.40^{\pm 0.26}$ \\\bottomrule
\end{tabular}}
\caption{Results on QALB-2014, QALB-2015 Test sets under Normalization Methods.}\label{tab: v2}
\end{table*}

%% file: Results_and_Discussion.tex
\section{Discussion}\label{D}

\noindent\textbf{LLMs and ChatGPT.} ChatGPT demonstrates remarkable ability to outperform other fully trained models by learning from only a few examples, particularly five-shot under both few-shot CoT and EP prompting strategies. Nevertheless, ChatGPT's performance lags behind AraT5\textsubscript{v2} and AraBART, suggesting potential areas for improvements in prompting strategies to fully exploit ChatGPT models. Models such as Vicuna-13B as well as those trained on multilingual datasets like  Bactrian-X$_{\textit{llama}}$-7B and  Bactrian-X$_{\textit{bloom}}$-7B, tend to perform better. However, these models fail to match ChatGPT's performance which reinforces ChatGPT's superiority in this domain.

\noindent\textbf{Seq2seq approach.}
Despite being smaller in size, pretrained Language Models (PLMs) often outperform LLMs, especially models specifically trained for Arabic tasks, such as AraT5\textsubscript{v2} and AraBART. In contrast, mT0 and mT5, both of which are multilingual models, are surpassed by ChatGPT when using both prompting strategies from 3-shot, but still outperform smaller LLMs such as LLaMA, Alpaca and Vicuna. Moreover, the results underscore the advantages of synthetic data for PLMs, as evidenced by the consistent improvement in scores with additional data.

\noindent\textbf{Seq2edit approach.}
These models exhibit high precision scores and relatively low recall scores, suggesting their strengths in making corrections rather than detecting errors. This trend can be explained by the absence of \textit{g-transformations}. For instance, in the case of English GECToR models, \textit{g-transformations} enable a variety of changes, such as case alterations and grammatical transformations. However, in this work we only rely on the 'merge' \textit{g-transformations} from the GECToR model as crafting effective \textit{g-transformations} for Arabic, a language with rich morphological features, poses significant challenges, limiting the model's ability to effectively detect errors. Developing specific \textit{g-transformations} for Arabic could significantly improve performance in these models.

\noindent\textbf{Data augmentation.}
Data augmentation results underscore the potential of synthetic data, generated by ChatGPT, in enhancing model performance. Our findings reveal that not just the quantity, but the quality of synthetic data, is crucial for achieving optimal performance. The relative underperformance of models further trained with synthetically generated data examples emphasizes this conclusion. Improvements we observe when expanding the dataset from 5M to 10M and from 10M to 11M are similar, even though the quantity of additional data vary. This can be attributed to the quality of the sources as the data for 5M and 10M were derived from noisier online commentaries, while the 11M data was derived from the OSIAN corpus~\cite{zeroual-etal-2019-osian}. Furthermore, our results on decoding methods on scaled datasets indicate that the chosen method can significantly influence precision and recall, emphasizing the need to select the right method depending on the specific task at hand.

\noindent\textbf{Best model in comparison.}
Although our main objective is not to develop the best model for AGEC, our AraT5\textsubscript{v2} (11M) model as detailed in Table~\ref{tab: normalization_2} excels in comparison to previous SOTA ($71.82$ vs. $72.90$). It is worth noting that contemporaneous work by~\citet{alhafni2023advancements}  introduces a new alignment algorithm that is much better than that employed by the shared task evaluation code we use. They also present an AGEC model. In personal communication with the authors, they confirmed their alignment algorithm through which we can perform direct and fair comparisons, and the data split on ZAEBUC dataset~\cite{habash-palfreyman-2022-zaebuc} will be released once their work is published through peer-review. Different from their work, our models are also dependency-free. For example, we do not exploit any morphological analyzers.
\input{Tables/normalization_table}

%% file: Tables/normalization_table.tex
\begin{table}[t]
\centering
\scriptsize
\renewcommand{\arraystretch}{1.35}
\resizebox{\columnwidth}{!}{%
\begin{tabular}{llrrrr}\toprule
\multirow{2}{*}{\textbf{Test Set}} &\multirow{2}{*}{\textbf{~~~~~~~~~~~~~~Models}} &\multicolumn{4}{c}{\textbf{Exact Match}} \\\cmidrule{3-6}
& &\textbf{P} &\textbf{R} &\textbf{F\textsubscript{1.0}} &\textbf{F\textsubscript{0.5}} \\\midrule
\multirow{6}{*}{\textbf{QALB-2014}}
&\citet{SOLYMAN2021303} &$\textbf{79.06}$ &$65.79$ &$71.82 $&$\textbf{75.99}$ \\
& \citet{mohit2014first} &$73.34$ &$63.23$ &$67.91$ &$71.07 $\\
&GPT4 (5-shot) &$69.46$ &$61.96$ &$65.49 $&$67.82$ \\
&ARBERT\textsubscript{v2} (3-step) & $74.17^{\pm 0.22}$ & $47.34^{\pm 0.30}$ & $57.79^{\pm 0.29}$ & $66.62^{\pm 0.26}$ \\
&AraT5\textsubscript{v2} (11m) & $76.94\textsuperscript{$\pm$0.67}$ & $\textbf{69.26}\textsuperscript{$\pm$0.73}$ &$ \textbf{72.90}\textsuperscript{$\pm$0.68}$ & $75.27\textsuperscript{$\pm$0.67}$ \\
\cmidrule{2-6}
\multirow{6}{*}{\textbf{QALB-2015}}
&\citet{SOLYMAN2021303} &$\textbf{80.23}$ &$63.59$ &$70.91$ &$\textbf{76.24}$ \\
&\citet{rozovskaya2015second}  &$88.85$ &$61.76$ &$72.87$ &$81.68$ \\
&ChatGPT (3-shot) + EP &$52.33$ &$47.57$ &$49.83$ &$54.10$ \\
&ARBERT\textsubscript{v2} (3-step) & $73.92^{\pm 0.28}$ & $53.15^{\pm 0.59}$ & $61.84^{\pm 0.49}$ & $68.56^{\pm 0.39}$ \\
&AraT5\textsubscript{v2} (11m) & $72.10^{\pm 0.31}$ &$ \textbf{73.59}\textsuperscript{$\pm$0.70}$ & $\textbf{72.84}\textsuperscript{$\pm$0.40}$ & $72.40^{\pm 0.30}$ \\
\bottomrule
\end{tabular}}%
\caption{Results on QALB-2014, QALB-2015 Test sets compared to recent works.}\label{tab: normalization_2} 
\end{table}

%% file: Conclusion.tex
\section{Conclusion}\label{C}
This paper provided a detailed exploration of the potential of LLMs, with a particular emphasis on ChatGPT for AGEC. Our study highlights ChatGPT's promising capabilities, in low-resource scenarios, as evidenced by its competitive performance on few-shot setttings. However, AraT5\textsubscript{v2} and AraBART still exhibit superior results across various settings and error types. Our findings also emphasize the role of high-quality synthetic data, reinforcing that both quantity and quality matter in achieving optimal performance. Moreover, our work unveils trade-offs between precision and recall in relation to dataset size and throughout all the other experimental settings. These insight, again, could inform future strategies for improving GEC systems. Although our exploration of ChatGPT's performance on AGEC tasks showcases encouraging results, it also uncovers areas ripe for further study. Notably, there remains significant room for improvement in GEC systems, particularly within the context of low-resource languages. Future research  may include refining prompting strategies, enhancing synthetic data generation techniques, and addressing the complexities and rich morphological features inherent in the Arabic language. 

\section{Limitations}\label{sec:limit}
We identify the following limitations in this work:
\begin{enumerate}
\item This work is primarily focused on MSA and does not delve into dialectal Arabic (DA) or the classical variety of Arabic (CA). While there exist DA resources such as the MADAR corpus~\cite{bouamor-etal-2018-madar}, their primary application is for dialect identification (DID) and machine translation (MT), making them unsuitable for our specific AGEC objectives. A more comprehensive coverage could be achieved with the development and introduction of datasets specifically tailored for the dialects in question.

\item This work aimed to examine the potential of LLMs, with an emphasis on ChatGPT, by comparing them to fully pretrained models. However, uncertainty surrounding the extent of Arabic data on which ChatGPT has been trained, poses challenges for direct comparisons with other pretrained models. Additionally, LLMs are primarily fine-tuned for English-language data. While prior studies have demonstrated their effectiveness in other languages, the limited amount of pretraining data for non-English languages complicates a straightforward comparison.

\item The scope of this work is primarily centered on sentence-level GEC. This limitation arose due to the official ChatGPT API, at the time of our study, allowed a maximum of 4,097 tokens, making it unsuitable for longer texts and precluding document-level GEC tasks. However, it's worth noting that document-level correction, offers a broader context that's vital for addressing certain grammatical inconsistencies and errors~\cite{yuan-bryant-2021-document}. With the recent introduction of a newer API that accommodates extended texts, future endeavors can potentially address document-level GEC, utilizing datasets such as QALB-2015 L2 and the newly introduced ZAEBUC corpus.
\end{enumerate}

\section{Ethics Statement and Broad Impact}\label{sec:ethics}
\textbf{Encouraging research development and contributing to a collaborative research culture.} Progress in AGEC has been stagnant for a long time due to the lack of benchmark datasets. This can be attributed to the extensive time and cost involved in creating these datasets. As a result, advancing AGEC has proven challenging. With the recent development of LLMs and their capabilities, there is potential for these models to expedite the creation of datasets. By doing so, they can significantly reduce both time and cost, as has been observed in other languages. We hope our work will inspire further exploration into the capabilities of LLMs for AGEC, thus aiding in the progress of this field.

\noindent\textbf{Advancing Second Language Learning through LLMs.}
With increasing interest in second language learning, ensuring accuracy and effectiveness of written language has become significant for pedagogical tools. Nowadays, individuals treat LLMs as their own writing assistants. Therefore, LLMs in the context of educational applications and more specifically GEC is becoming increasingly important. As such, introducing works in the development of tools that aid assistance in writing can help bridge the gap between non-native speakers and fluent written communication, enhancing the efficacy of educational tools. Especially with Arabic, being a collection of a diverse array of languages and dialectal varieties, we hope this will inspire more work to ensure comprehensive coverage and improved support for all learners. However, it is crucial to emphasize the ethical implications of using AI-driven educational tools. It's essential that these tools remain unbiased, transparent, and considerate of individual learning differences, ensuring the trustworthiness and integrity of educational platforms for every learner.

\noindent\textbf{Data privacy.} In relation to the data used in this work, all datasets are publicly available. Therefore, we do not have  privacy concerns.

%% file: ack.tex
\section*{Acknowledgments}\label{sec:acknow}
We acknowledge support from Canada Research Chairs (CRC), the Natural Sciences and Engineering Research Council of Canada (NSERC; RGPIN-2018-04267), the Social Sciences and Humanities Research Council of Canada (SSHRC; 435-2018-0576; 895-2020-1004; 895-2021-1008), Canadian Foundation for Innovation (CFI; 37771), Digital Research Alliance of Canada,\footnote{\href{https://alliancecan.ca}{https://alliancecan.ca}} and UBC Advanced Research Computing-Sockeye.\footnote{\href{https://arc.ubc.ca/ubc-arc-sockeye}{https://arc.ubc.ca/ubc-arc-sockeye}}

%% file: Appendix.tex
\clearpage


\section{Related Works} \label{app:RW_approaches}

\paragraph{Sequence to sequence approach.} Transformer-based Language Models (LMs) have been integral to advancements in GEC. These models have substantially transformed the perception of GEC, reframing it as a MT task. In this framework, erroneous sentences are considered as the source language, and the corrected versions as the target language. This perspective, which has led to SOTA results in the CONLL 2013 and 2014 shared tasks \cite{bryant2022grammatical,ng-etal-2013-conll,ng-etal-2014-conll}, reinterprets GEC as a low-resource or mid-resource MT task. Building on this paradigm, \citet{junczys-dowmunt-etal-2018-approaching} successfully adopted techniques from low-resource NMT and Statistical Machine Translation (SMT)-based GEC methods, leading to considerable improvements on both the CONLL and JFLEG datasets.

\paragraph{Sequence tagging approach.} Sequence tagging methods, another successful route to GEC, are showcased by models like GECToR \cite{omelianchuk2020gector}, LaserTagger \cite{malmi2019encode}, and the Parallel Iterative Edit (PIE) model ~\cite{awasthi-etal-2019-parallel}. By viewing GEC as a text editing task, these models make edits predictions instead of tokens, label sequences rather than generating them, and iteratively refine predictions to tackle dependencies. Employing a limited set of output tags, these models apply edit operations on the input sequence, reconstructing the output. This technique not only capably mirrors a significant chunk of the target training data, but it also diminishes the vocabulary size and establishes the output length as the source text's word count. Consequently, it curtails the number of training examples necessary for model accuracy, which is particularly beneficial in settings with sparse human-labeled data ~\cite{awasthi-etal-2019-parallel}.

\paragraph{Instruction fine-tuning.} LLMs have revolutionized NLP, their vast data-learning capability enabling diverse task generalizations. Key to their enhancement has been instructional finetuning, which fortifies the model's directive response and mitigates the need for few-shot examples ~\cite{ouyang2022training,wei2022chain,sanh2021multitask}. A novel approach, Chain of Thought (CoT), directs LLMs through a series of natural language reasoning, generating superior outputs. Proven beneficial in 'Let’s think step by step' prompts ~\cite{wei2022chain}, CoT has harnessed LLMs for multi-task cognitive tasks ~\cite{kojima2022large} and achieved SOTA results in complex system-2 tasks like arithmetic and symbolic reasoning.

\paragraph{ChatGPT.} In the specific realm of GEC, LLMs have demonstrated its potential.  \citet{fang2023chatgpt} applied zero-shot and few-shot CoT settings using in-context learning for ChatGPT ~\cite{brown2020language} and evaluated its performance on three document-level English GEC test sets. Similarly, ~\citet{wu2023chatgpt} carried out an empirical study to assess the effectiveness of ChatGPT in GEC, in the CoNLL2014 benchmark dataset.

\paragraph{Development in AGEC} Arabic consists of a collection of diverse languages and dialectal varieties with Modern Standard Arabic (MSA) being the current standard variety used in government and pan-arab media as well as education~\cite{abdul-mageed-etal-2020-nadi}. The inherent ambiguity of Arabic at the orthographic, morphological, syntactic, and semantic levels makes AGEC particularly challenging. Optional use of diacritics further introduces orthographic ambiguity~\cite{belkebir2021automatic}, making AGEC even harder.

Despite these hurdles, progress has been made in AGEC. For dataset development, the QALB corpus~\cite{zaghouani-etal-2014-large} was utilized. During the QALB-2014 and 2015 shared tasks~\cite{mohit2014first, rozovskaya-etal-2015-second}, the first AGEC datasets containing comments and documents from both native (L1) and Arabic learner (L2) speakers were released. Furthermore, the more recent ZAEBUC corpus~\cite{habash-palfreyman-2022-zaebuc}, which features essays from first-year university students at Zayed University in the UAE, has also been released. There has also been work on generating synthetic data. \citet{SOLYMAN2021303, SOLYMAN2023101572} apply Convolutional neural network (CNN) to generate synthetic parallel data using unsupervised noise injection techniques showing improvements in the QALB-2014 and 2015 benchmark datasets. In terms of model development, ~\citet{watson2018utilizing} developed a character-level seq2seq model that achieved notable results on AGEC L1 data, marking prgoress from basic classifier models~\cite{rozovskaya-etal-2014-columbia} and statistical machine translation models~\cite{jeblee-etal-2014-cmuq}. More recently, ~\citet{solyman2022automatic,SOLYMAN2021303} introduced novel design that incorporates dynamic linear combinations and the EM routing algorithm within a seq2seq Transformer framework.

\section{Instruction Fine-tuning LLMs}\label{app:IF}
\subsection{Instructions for LLMs}
Instruction format used for training is provided in Table~\ref{tab:llamadata} and instructions used for training are shown in Table~\ref{tab:llamains}.
\input{Tables/llama_format}
\input{Tables/LLama_Ins}

\subsection{Baseline and experimental setup for LLMs and ChatGPT}\label{app:BES}
For LLMs, evaluation was only done on the QALB-2014 Test set, for two main reasons. First was due to the high cost in producing results using ChatGPT and we were able to observation of a similar trend in our preliminary experiment with ChatGPT-3.5 Turbo on the QALB-2015. Second, as instruction fine-tuned were predominantly compared against ChatGPT's performance, we also evaluate them only on the QALB-2014 Test set.  These Results can be found in Table~\ref{tab:ChatGPT_2015}.
\input{Tables/2015_chatgpt}

\section{Sequence Tagging Approach}\label{app:GECToR}
The training procedure detailed in the original GECToR paper~\cite{omelianchuk2020gector} encompasses three stages: 
\begin{enumerate}
    \item [\textbf{1.}] Pre-training on synthetically generated sentences with errors.
    \item [\textbf{2.}] Fine-tuning solely on sentences that contain errors.
    \item [\textbf{3.}] Further fine-tuning on a mix of sentences, both with and without errors. 
\end{enumerate}
For our training process, we pre-train the model on the complete AGEC dataset~\cite{SOLYMAN2021303}, use the reverseGold dataset for stage 2, and employ the gold training data in the third stage. Moreover, as some corrections in a sentence depend on others, applying edit sequences once may not be enough to correct the sentence fully. To address this issue, GECToR employs an iterative correction approach from ~\citet{awasthi-etal-2019-parallel}. However, in our experiments, we find that the iterative correction approach does not result in any tangible improvement. Therefore, we set our iterations to $3$.





\section{Normalization Methods}\label{app:errors_normaliszation}

\subsection{Normalization examples}\label{app:Error_type_examples}
Examples of text under each normalization methods can be found in Table~\ref{tab:errors_normaliszation}
\input{Tables/Errors_Examples_2}

\subsection{Arabic Learner Corpus error type taxonomy }\label{app:acl}
The ALC error type taxonomy can be found in Table~\ref{tab:errors_class}.
\input{Tables/Errors_Class}
\subsection{Hyperparameters}\label{app: Hyperparameters}
The Hyperparameters used for training are shown in Table \ref{tab:hyperparameters_all_models}.
\input{Tables/hypess}
\subsection{Dev results}\label{app: 2014 dev Results}
Results on the Dev set are presented in Table \ref{tab: dev}.
\input{Tables/dev_gec}
\subsection{ARETA results}\label{app: areta}
Full results evaluated using  ARETA  are presented in Table~\ref{tab:model_performance_comparison}.
\input{Tables/Areta_updated}

%% file: Tables/llama_format.tex
\begin{table*}[]
\centering
\scriptsize
\resizebox{0.90\columnwidth}{!}{
\begin{tabular}{lr}
\toprule
~~~~~~~~~ \textbf{Instruction fine-tune Format }\\
\toprule
\RL{ فيما يلي أمر توجيه يصف مهمة مرتبطة بمدخل لتزويد} \\
~~\RL{  النص بسياق اضافي. يرجى صياغة ردود مناسبة لتحقق} \\
~~~~~~~~~~~~~~~~~~~~~~~~~~~~~~~~~\RL{  الطلب بطريقة مناسبة و دقيقة.} \\
\colorbox{blue!25}{\#\#\# \RL{: الأمر/ التوجيه} }\\
\textbf{\textit{\RL{قم بتصحيح كل الأخطاء الكتابية في النص التالي:}}}\\
\colorbox{pink!55}{\#\#\#\RL{ : المدخل}} \\
\textbf{\textit{\< الفرس . > \textcolor{red}{ \< يرب > }    \< الرجل >}} \\
\colorbox{yellow!50}{\#\#\#\RL{ : الرد}} \\
\textbf{\textit{ \<  الفرس . > \textcolor{green}{  \< يركب > }    \< الرجل >}} \\
\bottomrule
\end{tabular}
}
\caption{Modified data format for the LLaMA instruction fine-tuning step.}\label{tab:llamadata}
\end{table*}

%% file: Tables/LLama_Ins.tex
\begin{table*}[h]
\centering
\renewcommand{\arraystretch}{1.95}
\resizebox{0.95\textwidth}{!}{
\begin{tabular}{lr}\toprule
\textbf{Translated in English} &\textbf{Instructions Samples} \\\midrule
Correct all written errors in the following text except for a thousand, ya and punctuation: &\RL{قم بتصحيح كل الأخطاء الكتابية في النص التالي ماعدا المتعلقة بالألف والياء وعلامات الترقيم:} \\
Please verify spelling, grammatical scrutiny, and correct all errors in the following sentence, except for punctuation: &\RL{الرجاء التدقيق الإملائي والتدقيق النحوي و تصحيح كل الأخطاء في الجملة التالية إلا الخاصة بعلامات الترقيم:} \\
Explore the grammatical errors and repair them except for punctuation marks such as a comma, or a question marks, etc: &\RL{قم بإستكشاف أخطاء التدقيق الإملائي وإصلاحها ماعدا المتعلقة بعلامات الترقيم كالفاصلة  أو علامة إستفهام ، إلخ:} \\
Can you correct all errors in the following text except those related to punctuation such as commas, periods, etc: &\RL{هل يمكنك كل الأخطاء الموجودة في النص التالي ماعدا المتعلقة بعلامات الترقيم كالفاصلة ، النقطة ، إلخ : } \\
Can you fix all spelling and grammatical errors, except for the mistakes of the "Alif" and "Ya": &\RL{هل يمكنك إصلاح كل الأخطاء الإملائية والنحوية ماعدا الأخطاء الخاصة بالألف والياء:} \\
Please explore the grammatical spelling errors and repair them all, except for the mistakes related to the "Alif" and "Ya" &\RL{الرجاء إستكشاف أخطاء التدقيق الإملائي النحوي وإصلاحها كلها ماعدا الأخطاء المتعلقة بالألف والياء:} \\
Correct all the written errors in the following text except for the "Alif" and "Ya": &\RL{قم بتصحيح كل الأخطاء الكتابية في النص التالي ماعدا المتعلقة بالألف والياء:} \\
Please correct all errors in the following sentence: &\RL{الرجاء تصحيح كل الأخطاء الموجودة في الجملة التالية:} \\
\bottomrule
\end{tabular}}
\caption{Different instructions used for instruction fine-tuning.}\label{tab:llamains}
\end{table*}

%% file: Tables/2015_chatgpt.tex
\begin{table*}[t]
\centering
\resizebox{\columnwidth}{!}{%
\begin{tabular}{llcccc}\toprule
\multirow{2}{*}{\textbf{Settings}} &\multirow{2}{*}{\textbf{Models}} &\multicolumn{4}{c}{\textbf{Exact Match}} \\\cmidrule{3-6}
& &\textbf{P} &\textbf{R} &\textbf{F\textsubscript{1.0}} &\textbf{F\textsubscript{0.5}} \\ \toprule 
\multirow{2}{*}{\textbf{+ CoT}} 
&ChatGPT (3-shot) &$49.89$ &$46.72$ &$48.22$ &$49.49$ \\
&ChatGPT (5-shot) &$52.33$ &$47.57$ &$49.83$ &$51.15$ \\
\toprule
\end{tabular}%
}
\caption{Performance of ChatGPT-3.5 on QALB-2015 Test set.} 
\label{tab:ChatGPT_2015}
\end{table*}

%% file: Tables/Errors_Examples_2.tex
\begin{table*}[t]
\centering
\renewcommand{\arraystretch}{1.5}
 \resizebox{\linewidth}{!}{%

\begin{tabular}{lr}

\toprule
\textbf{Normalisation Method}             & \multicolumn{1}{c}{\textbf{Example}}                                                                                                                                                                                                               \\ \toprule
\textbf{Normal}                 &

\RL{نحن معشر العرب نعرف إلا الشماتة ، ولكن يجب أن ندرس هذه الحالة ونحن المخرج منها من الاقتصاد الإسلامي.}

\\
\textbf{No Alif/Ya}              & 
\RL{ نحن معشر العرب نعرف الا الشماتة ، ولكن يجب ان ندرس هذه الحالة ونحن المخرج منها من الاقتصاد الاسلامي.} 
\\

\textbf{No Punct}                &

\RL{نحن معشر العرب نعرف إلا الشماتة ولكن يجب أن ندرس هذه الحالة ونحن المخرج منها من الاقتصاد الإسلامي} 

\\
\textbf{No Alif/Ya \& Punct}  & 
\RL{نحن معشر العرب نعرف الا الشماتة ولكن يجب ان ندرس هذه الحالة ونحن المخرج منها من الاقتصاد الاسلامي}
\\ \toprule
\end{tabular}
}
\caption{Examples of normalized text: with Alif/Ya errors removed, punctuation removed, and both Alif/Ya errors and punctuation removed.  \label{tab:errors_normaliszation}}
\end{table*}

%% file: Tables/Errors_Class.tex
\begin{table*}[t]
\centering
\renewcommand{\arraystretch}{1.}
\resizebox{0.85\textwidth}{!}{%
\begin{tabular}{ccl}
\toprule
\textbf{Class}                          & \textbf{Sub-class} & \textbf{Description}                                \\\toprule
\multirow{12}{*}{\textbf{Orthographic}} & \textbf{OH}        & \textbf{Hamza error}                                \\
                                        & \textbf{OT}        & Confusion in Ha and Ta Mutadarrifatin               \\
                                        & \textbf{OA}        & \textbf{Confusuion in Alif and Ya Mutadarrifatin}   \\
                                        & \textbf{OW}        & Confusion in Alif Fariqa                            \\
                                        & \textbf{ON}        & Confusion Between Nun and Tanwin                    \\
                                        & \textbf{OS}        & Shortening the long vowels                          \\
                                        & \textbf{OG}        & Lengthening the short vowels                        \\
                                        & \textbf{OC}        & Wrong order of word characters                      \\
                                        & \textbf{OR}        & Replacement in word character(s)                    \\
                                        & \textbf{OD}        & Additional character(s)                             \\
                                        & \textbf{OM}        & Missing character(s)                                \\
                                        & \textbf{OO}        & Other orthographic errors                           \\ \midrule
\multirow{9}{*}{\textbf{Morphological}} & \textbf{MI}        & Word inflection                                     \\
                                        & \textbf{MT}        & Verb tense                                          \\
                                        & \textbf{MO}        & Other morphological errors                          \\
                                        & \textbf{XF}        & Definiteness                                        \\
                                        & \textbf{XG}        & Gender                                              \\
                                        & \textbf{XN}        & Number                                              \\
                                        & \textbf{XT}        & Unnecessary word                                    \\
                                        & \textbf{XM}        & Missing word                                        \\
                                        & \textbf{XO}        & Other syntactic errors                              \\ \midrule
\multirow{3}{*}{\textbf{Semantic}}      & \textbf{SW}        & Word selection error                                \\
                                        & \textbf{SF}        & Fasl wa wasl (confusion in conjunction use/non-use) \\
                                        & \textbf{SO}        & Other semantic errors                               \\ \midrule
\multirow{4}{*}{\textbf{Punctuation}}   & \textbf{PC}        & Punctuation confusion                               \\
                                        & \textbf{PT}        & Unnecessary punctuation                             \\
                                        & \textbf{PM}        & Missing punctuation                                 \\
                                        & \textbf{PO}        & Other errors in punctuation                         \\
\textbf{Merge}                          & \textbf{MG}        & Words are merged                                    \\\midrule
\textbf{Split}                          & \textbf{SP}        & Words are split        \\ 
\toprule
\end{tabular}}

\caption{ The ALC error type taxonomy extended with merge and split classes} \label{tab:errors_class}
\end{table*}

%% file: Tables/hypess.tex
\begin{table*}[!htp]\centering
\scriptsize
\resizebox{\textwidth}{!}{%
\begin{tabular}{lccc}\toprule
\textbf{Hyperparameter} & \textbf{Seq2seq} & \textbf{Decoder Only (LLMs)} & \textbf{Seq2Edit Encoder Only} \\\toprule
Learning Rate & $5 \times 10^{-5}$ & $2 \times 10^{-5}$ & $1 \times 10^{-5}$ \\
Train Batch Size & 4 & 8 & 8 \\
Eval Batch Size & 4 & 8 & 8 \\
Seed & 42 & 42 & 42 \\
Gradient Accumulation Steps & 8 & 8 & 8 \\
Total Train Batch Size & 32 & 64 & 64 \\
Optimizer & Adam (betas=(0.9,0.999), epsilon=$1 \times 10^{-8}$) & AdamW (betas=(0.9,0.999), epsilon=$1 \times 10^{-7}$) &  AdamW (betas=(0.9,0.999), epsilon=$1\times 10^{-8}$)  \\
LR Scheduler Type & Cosine & Linear & Cosine \\
Num Epochs & 50 & 4 & 100 \\ \bottomrule
\end{tabular}}
\caption{Summary of hyperparameters used for model training.}\label{tab:hyperparameters_all_models}
\end{table*}

%% file: Tables/dev_gec.tex
\begin{table*}[!htp]\centering
\scriptsize
\resizebox{\textwidth}{!}{
\begin{tabular}{lrrrrr|rrrr|rrrr|rrrrr}\toprule
\multirow{2}{*}{\textbf{Settings}} &\multirow{2}{*}{\textbf{Models}} &\multicolumn{4}{c}{\textbf{Exact Match}} &\multicolumn{4}{c}{\textbf{No Alif / Ya Errors}} &\multicolumn{4}{c}{\textbf{No Punctuation}} &\multicolumn{4}{c}{\textbf{No Puncation and Alif / Ya Errors}} \\\cmidrule{3-18}
& &\textbf{P} &\textbf{R} &\textbf{F\textsubscript{1.0}} &\textbf{F\textsubscript{0.5}} &\textbf{P} &\textbf{R} &\textbf{F\textsubscript{1.0}} &\textbf{F\textsubscript{0.5}} &\textbf{P} &\textbf{R} &\textbf{F\textsubscript{1.0}} &\textbf{F\textsubscript{0.5}} &\textbf{P} &\textbf{R} &\textbf{F\textsubscript{1.0}} &\textbf{F\textsubscript{0.5}} \\\midrule
\multirow{4}{*}{\textbf{Seq2Edit}} &ARBERTv2 &73.30 &47.85 &57.90 &66.25 &65.60 &44.20 &52.81 &59.81 &72.38 &48.75 &58.26 &65.98 &57.40 &33.90 &42.63 &50.41 \\
&ARBERT\textsubscript{v2} 3-stage &74.65 &46.70 &57.46 &66.67 &65.00 &41.20 &50.43 &58.27 &75.50 &44.50 &56.00 &66.27 &55.70 &27.50 &36.82 &46.22 \\
&MARBERT\textsubscript{v2} &72.95 &47.65 &57.65 &65.95 &64.60 &43.20 &51.78 &58.78 &73.72 &44.16 &55.23 &65.02 &56.80 &34.20 &42.69 &50.17 \\
&MARBERT\textsubscript{v2} 3-stage &74.55 &45.75 &56.70 &66.21 &65.10 &41.30 &50.54 &58.37 &75.41 &45.52 &56.77 &66.66 &56.00 &29.20 &38.38 &47.31 \\\hline 
\multirow{5}{*}{\textbf{LLMs}} &LLama-7B &58.20 &32.50 &41.71 &50.25 &35.50 &16.70 &22.71 &28.98 &19.60 &54.30 &28.80 &22.47 &65.10 &32.00 &42.91 &53.94 \\
&Alpaca-7B &42.20 &31.20 &35.88 &39.42 &42.20 &33.40 &37.29 &40.09 &82.20 &62.20 &70.81 &77.23 &62.20 &39.50 &48.32 &55.79 \\
&Vicuna-13B &63.90 &51.00 &56.73 &60.82 &51.40 &39.30 &44.54 &48.42 &83.90 &73.90 &78.58 &81.69 &68.50 &49.00 &57.13 &63.45 \\
&Bactrian-X$_{\textit{bloom}}$-7B &60.80 &43.80 &50.92 &56.42 &53.70 &41.00 &46.50 &50.57 &79.40 &63.00 &70.26 &75.47 &62.00 &51.00 &55.96 &59.44 \\
&Bactrian-X$_{\textit{llama}}$-7B &58.60 &41.40 &48.52 &54.10 &51.00 &38.10 &43.62 &47.77 &77.00 &59.20 &66.94 &72.63 &58.60 &48.10 &52.83 &56.15 \\\hline
\multirow{7}{*}{\textbf{Seq2Seq}} &mT0 &69.35 &54.29 &60.90 &65.70 &57.45 &42.50 &48.86 &53.67 &82.35 &75.34 &78.69 &80.85 &70.20 &50.30 &58.61 &65.05 \\
&mT5 &69.00 &53.20 &60.08 &65.13 &56.70 &39.50 &46.56 &52.16 &81.00 &70.00 &75.10 &78.53 &68.00 &48.00 &56.28 &62.77 \\
&AraBART &72.00 &61.50 &66.34 &69.62 &60.00 &49.70 &54.37 &57.61 &85.00 &78.50 &81.62 &83.62 &74.00 &60.50 &66.57 &70.84 \\
&AraT5\textsubscript{v2} &74.50 &64.50 &69.14 &72.26 &63.50 &52.70 &57.60 &61.00 &88.00 &84.50 &86.21 &87.28 &81.50 &69.50 &75.02 &78.78 \\
&AraT5\textsubscript{v2} (5M) &75.33 &67.44 &71.17 &73.61 &64.55 &51.55 &57.32 &61.45 &89.22 &83.40 &86.21 &87.99 &81.30 &70.24 &75.37 &78.82 \\
&AraT5\textsubscript{v2} (10M) &75.90 &68.33 &71.92 &74.25 &65.34 &52.44 &58.18 &62.28 &89.88 &84.22 &86.96 &88.69 &82.34 &71.44 &76.50 &79.90 \\
&AraT5\textsubscript{v2} (11M) &77.85 &68.90 &73.10 &75.88 &66.33 &55.20 &60.26 &63.76 &90.10 &85.21 &87.59 &89.08 &84.55 &71.50 &77.48 &81.57 \\
\bottomrule
\end{tabular}}
\caption{Dev Set results on the QALB-2014 benchmark dataset.}\label{tab: dev}
\end{table*}

%% file: Tables/Areta_updated.tex
\begin{table*}[!htp]\centering
\scriptsize
\resizebox{\textwidth}{!}{
\begin{tabular}{cccccccc}\toprule
\textbf{CLASS} & \textbf{GECToR\_ARBERT} & \textbf{five-shot\_2014\_expertprompt} & \textbf{five-shot\_2014-chatgpt4} & \textbf{AraT5 (11M)} & \textbf{COUNT} \\\midrule
OH & 73.73 & 89.80 & 92.91 & 87.34 & 4902 \\
OT & 76.59 & 94.12 & 95.58 & 90.84 & 708 \\
OA & 78.63 & 84.66 & 88.93 & 87.35 & 275 \\
OW & 38.57 & 80.79 & 86.96 & 83.70 & 107 \\
ON & 0.00 & 0.00 & 0.00 & 0.00 & 0 \\
OG & 48.00 & 55.74 & 63.64 & 90.32 & 34 \\
OC & 21.43 & 28.57 & 53.66 & 87.18 & 22 \\
OR & 38.24 & 53.02 & 65.96 & 77.10 & 528 \\
OD & 33.76 & 51.89 & 59.60 & 73.07 & 321 \\
OM & 41.80 & 44.53 & 57.35 & 86.44 & 393 \\
OO & 0.00 & 0.00 & 0.00 & 0.00 & 0 \\
MI & 11.02 & 13.25 & 20.53 & 75.00 & 83 \\
MT & 0.00 & 7.84 & 11.43 & 62.50 & 7 \\
XC & 32.95 & 46.10 & 50.78 & 88.35 & 526 \\
XF & 6.06 & 17.98 & 23.81 & 76.92 & 29 \\
XG & 37.10 & 19.57 & 31.35 & 89.47 & 79 \\
XN & 25.19 & 25.79 & 31.25 & 88.12 & 108 \\
XT & 3.95 & 3.78 & 5.48 & 2.48 & 66 \\
XM & 2.04 & 4.14 & 6.38 & 1.07 & 26 \\
XO & 0.00 & 0.00 & 0.00 & 0.00 & 0 \\
SW & 50.51 & 21.25 & 33.38 & 8.29 & 219 \\
SF & 0.00 & 6.67 & 3.45 & 57.14 & 3 \\
PC & 60.89 & 56.25 & 47.59 & 74.98 & 713 \\
PT & 29.62 & 29.58 & 21.40 & 57.42 & 480 \\
PM & 55.24 & 54.21 & 52.09 & 67.08 & 5599 \\
MG & 25.05 & 75.96 & 79.70 & 64.80 & 434 \\
SP & 42.27 & 90.93 & 91.61 & 86.70 & 805 \\\midrule
\textbf{micro avg} & 55.67 & 60.05 & 64.51 & 57.28 & 16467 \\
\textbf{macro avg} & 30.84 & 39.13 & 43.51 & 61.62 & 16467 \\
\textbf{weighted avg} & 56.98 & 66.96 & 68.24 & 76.35 & 16467 \\
\bottomrule
\end{tabular}}
\caption{ Analysis of Error Type performances on the QALB-2014 Test set.}\label{tab:model_performance_comparison}
\end{table*}